\documentclass[10pt,twocolumn,letterpaper]{article}

\usepackage[final]{cvpr}
\usepackage{graphicx}
\usepackage{amsmath}
\usepackage{amssymb}
\usepackage{booktabs}
\usepackage{mathtools}
\usepackage[accsupp]{axessibility}

\makeatletter
\def\maketag@@@#1{\hbox{\m@th\normalfont\normalsize#1}}
\makeatother

\usepackage[pagebackref,breaklinks,colorlinks]{hyperref}

\usepackage{booktabs}       % professional-quality tables

\usepackage{times}
\usepackage{epsfig}
\usepackage{graphicx}
\usepackage{amsmath}
\usepackage{amssymb}
\usepackage{caption}
\usepackage{subcaption}
\usepackage{amsthm}

\usepackage{soul}
\usepackage{tabularx}

\usepackage{algorithm}
\usepackage{algorithmic}

\theoremstyle{definition}
\newtheorem{problem}{Problem}[section]
\theoremstyle{definition}
\newtheorem{remark}{Remark}[section]

\newcommand\blfootnote[1]{%
  \begingroup
  \renewcommand\thefootnote{}\footnote{#1}%
  \addtocounter{footnote}{-1}%
  \endgroup
}

\usepackage[capitalize]{cleveref}
\crefname{section}{Sec.}{Secs.}
\Crefname{section}{Section}{Sections}
\Crefname{table}{Table}{Tables}
\crefname{table}{Tab.}{Tabs.}

\def\cvprPaperID{2206} % *** Enter the CVPR Paper ID here
\def\confName{CVPR}
\def\confYear{2022}

\begin{document}

\title{Towards Unsupervised Domain Generalization}

% \and
\author{Xingxuan Zhang$^{\dag}$, Linjun Zhou$^{\dag}$, Renzhe Xu, Peng Cui*, Zheyan Shen, Haoxin Liu \\
Department of Computer Science, Tsinghua University\\
{\tt\small {xingxuanzhang@hotmail.com, zhoulj16@mails.tsinghua.edu.cn, 
xrz199721@gmail.com}} \\
{\tt\small {cuip@tsinghua.edu.cn,  shenzy17@mails.tsinghua.edu.cn, 1132462715@qq.com}}}
\maketitle

\begin{abstract}
  Domain generalization (DG) aims to help models trained on a set of source domains generalize better on unseen target domains. The performances of current DG methods largely rely on sufficient labeled data, which are usually costly or unavailable, however. Since unlabeled data are far more accessible, we seek to explore how unsupervised learning can help deep models generalize across domains. Specifically, we study a novel generalization problem called unsupervised domain generalization (UDG), which aims to learn generalizable models with unlabeled data and analyze the effects of pre-training on DG. 
  In UDG, models are pretrained with unlabeled data from various source domains before being trained on labeled source data and eventually tested on unseen target domains.
  Then we propose a method named Domain-Aware Representation LearnING (DARLING) to cope with the significant and misleading heterogeneity within unlabeled pretraining data and severe distribution shifts between source and target data. 
  Surprisingly we observe that DARLING can not only counterbalance the scarcity of labeled data but also further strengthen the generalization ability of models when the labeled data are insufficient. 
  As a pretraining approach, DARLING shows superior or comparable performance compared with ImageNet pretraining protocol even when the available data are unlabeled and of a vastly smaller amount compared to ImageNet, which may shed light on improving generalization with large-scale unlabeled data.
%   Extensive experiments clearly demonstrate the effectiveness of our method compared with state-of-the-art unsupervised learning counterparts. 
  
%   significantly improves the generalization ability of models not only with scarce labeled data, 
  
%   Surprisingly we observe the superior of unsupervised learned models to ImageNet pretrained ones even when the unlabeled data for unsupervised training is of much smaller amount compared to ImageNet. To further improve the generalization ability, we propose a novel domain-irrelevant representation learning method and quantify the margin between our method and current contrastive learning methods.
%   For sensible evaluation of our method and other unsupervised learning method, three novel experimental settings of unsupervised DG are proposed. The experimental results on these settings clearly demonstrate the effectiveness of our method compared with state-of-the-art counterparts.
\end{abstract}

\blfootnote{$\dag$Equal contribution, *Corresponding author}

\section{Introduction}
Deep neural network based approaches have achieved striking performance in tasks where training and test data share similar distribution \cite{he2016deep,he2020momentum}.
However, under considerable distribution shifts, they can significantly fail \cite{bengio2019meta, engstrom2019exploring, hendrycks2018benchmarking, recht2019imagenet, su2019one}. To address this problem, the literature in domain generalization (DG) proposes algorithms that have access to labeled data from multiple domains or environments during training and generalize well to unseen test domains \cite{ghifary2015domain, khosla2012undoing, li2017deeper, li2019episodic, muandet2013domain, wang2020learning}.

Sufficient labeled data are crucial for current DG methods to learn domain invariant features, which are proved to be generalizable to unseen domains \cite{muandet2013domain, albuquerque2020adversarial, shao2019multi, wang2020unseen}. A common and effective approach to learning discriminative features in DG is to enlarge the available data space with augmentations of source domains \cite{carlucci2019domain, zhou2020deep, zhou2020learning}. With sufficient source data and strong augmentations, even empirical risk minimization (ERM) can outperform previous state-of-the-art methods \cite{gulrajani2020search}. Nevertheless, both augmentation-based methods and carefully hyperparameter tuned ERM assume adequate labeled data from multiple domains available for representation learning. 

As manually labeled data can be costly or unavailable while unlabeled data are far more accessible, we study a novel generalization problem called unsupervised domain generalization (UDG). UDG aims to unsupervised learn discriminative representations that generalize well across domains and thus  reduce the dependence of DG methods on labeled data. Specifically, models are trained with unlabeled heterogeneous data before finetuned and evaluated on labeled data, so that methods for UDG can be easily assembled with current DG methods as pretraining and study how pre-training affects models' generalization ability. 

In the field of unsupervised learning \cite{harwood2017smart, saunshi2019theoretical, wu2017sampling}, contrastive learning (CL) advances in discriminative representation learning for downstream tasks compared to its counterparts \cite{he2020momentum, tian2019contrastive, chen2020simple}. Actually, the objective of CL, which is to maximize the similarity between a given image and its variant under disturbance while contrasting with negatives \cite{fan2021does, wu2021contrastive, li2021dual}, agrees with the target of DG.
However, current CL only learns robust representations against predefined perturbation under independent and identically distributed (I.I.D) hypothesis \cite{bachman2019learning, henaff2020data, hjelm2018learning} and fails to consider severe distribution shifts across domains beyond predefined perturbation types \cite{oord2018representation, wu2018unsupervised}. With 
samples from various domains as negative pairs, current CL methods leverage both domain-related (\textit{i.e.}, features irrelevant to categories) and category-discriminative features to push their representations away.
Furthermore, in UDG, the distribution shifts across domains in training data are significant and can not be fully counterweighed via data transformations (for instance, one can hardly transform a dog in sketch to photo). The strong heterogeneity induces models to leverage the domain-related features to distinguish one sample from its negatives \cite{arjovsky2019invariant, shah2020pitfalls} and thus, hinders the learning of an invariant representation space where dissimilarity across domains is minimized \cite{matsuura2020domain, muandet2013domain}.
Thus current contrastive learning can not perfectly handle the UDG problem.

To address this problem, we propose Domain-Aware Representation LearnING (DARLING), a novel contrastive learning algorithm for UDG which unifies objectives of DG and contrastive learning. 
To force the model to ignore domain-related features, we select valid sources of negative samples for any given queue according to the similarity between different domains. Specifically, the more similar two domains are, the more likely two samples in a negative pair are selected from these two domains, respectively. 
Intuitively, consider samples from two domains with enormous differences in distribution, the domain-related features of which are discriminative enough to distinguish them from each other and, in turn, boost variance across domains in the representation space.
On the contrary, if a negative pair of samples comes from a single domain and shares identical domain-related features, domain-irrelevant representations are learned to contrast them.

As shown in Section \ref{exp}, the proposed unsupervised pretraining protocol achieves a significant improvement in generalization even with raw ERM finetuning, indicating that the UDG problem gives an effective and enlightening complementary to supervised methods for DG. 
We further show that DARLING outperforms state-of-the-art counterparts by a considerable margin with quantitative and qualitative experiments. 
Moreover, prepositive unsupervised learning can be considered as a protocol of pretraining. Although initialization of weights pretrained on ImageNet shows unparalleled effectiveness on independent and identically distributed (I.I.D.) tasks, we argue that it lacks rationality for the DG problem. Since ImageNet can be considered as a set of data sampled from latent domains \cite{shen2021towards, zhang2021deep}, the distribution shifts across domains are not as significant as most DG datasets \cite{he2021towards,li2017deeper,peng2019moment}, resulting in insufficient heterogeneity for models to learn a generalizable representation space. Thus the protocol of unsupervised pretraining on heterogeneous unlabeled data is a reasonable alternative to initialization with weights pretrained on ImageNet for DG.

\section{Related work}
\noindent \textbf{Domain Generalization}. Domain generalization (DG) considers the generalization ability to novel domains of deep models trained on source domains where the heterogeneity caused by domain shifts is significant \cite{ghifary2015domain,khosla2012undoing}. 
A common approach is extracting domain-invariant features over multiple source 
domains \cite{dou2019domain,hu2020domain,li2018learning, li2018domain, motiian2017unified, piratla2020efficient, seo2019learning, zhang2021deep,xu2021stable,zhang2022towards}
or aggregating domain-specific modules \cite{mancini2018best, mancini2018robust} to conduct domain-invariant or domain-specific. 
Many works propose to enlarge the available data space with augmentation of source domains \cite{carlucci2019domain, volpi2018generalizing, qiao2020learning, shankar2018generalizing, zhou2020deep, zhou2020learning}. 
There are several approaches that exploit regularization with meta-learning \cite{dou2019domain, li2019episodic} and Invariant Risk Minimization (IRM) framework \cite{arjovsky2019invariant} for DG.
Despite the promising results achieved by current DG methods, all of them assume that the training data provide ample heterogeneity and knowledge for target categories. Such assumptions hinder DG methods from real applications. 

\noindent \textbf{Unsupervised learning}. Unsupervised representation learning generally involve two categories, namely generative and discriminative \cite{chen2020improved, doersch2015unsupervised}. Many of generative approaches rely on auto-encoder \cite{vincent2008extracting} or adversarial learning \cite{goodfellow2014generative}, where data and representations are jointly modeled \cite{donahue2016adversarial, dumoulin2016adversarially, gan2020large}. There are some self learning methods relying on auxiliary handcrafted prediction tasks such as image jigsaw puzzle\cite{noroozi2016unsupervised} and geometric transformations\cite{dosovitskiy2014discriminative} to learn representations.
Among discriminative method, Contrastive loss based approaches forces representation of different views of the same image closer with spreads representations of views from different images apart and achieves current state-of-the-art performance \cite{chen2020simple, grill2020bootstrap, he2020momentum, henaff2020data, hu2020adco, oord2018representation, tian2020makes} . 
As designed for problems under the I.I.D assumption, current contrastive learning can not distinguish domain or category related features, resulting in low training efficiency or misleading by spurious correlations between categories and domains. 

\section{Methods}
\subsection{Unsupervised Domain Generalization}
\paragraph{Notations} Let $\mathcal{X}$ be the feature space, $\mathcal{Y}$ the category label space, and $\mathcal{D}$ the domain label space. Accordingly, we use $X$, $Y$, $D$ to denote the random variables which take values in $\mathcal{X}$, $\mathcal{Y}$ and $\mathcal{D}$. A dataset $S$ contains $N_S$ samples $\{X_i, y_i, d_i\}_{i=1}^{N_S}$ sampled from a joint distribution $P^S$ on $\mathcal{X} \times \mathcal{Y} \times \mathcal{D}$. Let $P^S_X$, $P^S_Y$ and $P^S_D$ denote the marginal distribution of $P^S$ on $X$, $Y$, $D$ respectively. Let $\mathrm{Supp}(\cdot)$ denote the support of a distribution. For example, $\mathrm{Supp}(P^S_X)$ denotes the support of distribution $P^S_X$. Let $[K]$ denote the set $\{1, 2, \dots, K\}$.

We describe datasets as labeled when the category labels are available while others as unlabeled. We aim to learn a model generalizable to any unknown testing distribution. Formally, the problem is defined as follows:

\begin{problem} [Unsupervised Domain Generalization (UDG)] Let $S_\mathrm{UL} = \{X_i, d_i\}_{i=1}^{N_\mathrm{UL}}$ be the unlabeled dataset with $N_\mathrm{UL}$ samples from distribution $P^{S_\mathrm{UL}}$ and $S_\mathrm{L} = \{X_i, y_i\}_{i=1}^{N_\mathrm{L}}$ be the labeled dataset with $N_\mathrm{L}$ samples from distribution $P^{S_\mathrm{L}}$. There exists an unknown testing distribution $P^{S_\mathrm{test}}$ that satisfies
\small{
\begin{align}
    & \mathrm{Supp}\left(P^{S_\mathrm{test}}_D\right) \cap \left(\mathrm{Supp}\left(P^{S_\mathrm{UL}}_D\right) \cup 
    \mathrm{Supp}\left(P^{S_\mathrm{L}}_D\right)\right) = \emptyset. \label{eq:setting_domain}\\
    & \mathrm{Supp}\left(P^{S_\mathrm{test}}_Y\right) = \mathrm{Supp}\left(P^{S_\mathrm{L}}_Y\right) \label{eq:setting_label}.
\end{align}
  }
Given $S_\mathrm{UL}$, $S_\mathrm{L}$ and a loss function $\ell(X, Y; \theta)$, we aim to learn a model with parameters $\theta^*$ that achieves best performance on $P^{S_\mathrm{test}}$.
\begin{equation}
    \theta^* = \underset{\theta}{\arg \min} \, \mathbb{E}_{(X, Y, D) \sim P^{S_\mathrm{test}}} \left[\ell(X, Y; \theta)\right].   
\end{equation}

\end{problem}

\begin{remark} [Explanation of the two constraints]
    Following the standard DG setting, Equation \ref{eq:setting_domain} requires that there is no domain overlap between testing and all available training datasets, including labeled and unlabeled ones. Meanwhile, Equation \ref{eq:setting_label} requires that the category space should be the same between the testing dataset and the labeled dataset.
\end{remark}

This setting is sound since we can consider the source or mechanism of data generation as the domain while the latent structure of data other than domains determines the categories. Accordingly, the domain label is significantly easier to access while category labeling can be expensive, leading to a large scale of data with domain label while without category label.

\paragraph{UDG settings}
We specifically describe all possible 4 settings that support unsupervised domain generalization (UDG) according to the intersections in the category and domain spaces between unlabeled $P^{S_\mathrm{UL}}$ and labeled $P^{S_\mathrm{L}}$ data, namely \textit{all correlated}, \textit{domain correlated}, \textit{category correlated}, \textit{uncorrelated}.

\noindent \textbf{All correlated} When the data are partially and randomly labeled, the unlabeled and labeled data are homologous so that there can be overlap in the category and domain spaces between them.
Formally, $\mathrm{Supp}(P^{S_\mathrm{UL}}_D) = \mathrm{Supp}(P^{S_\mathrm{L}}_D)$, $\mathrm{Supp}(P^{S_\mathrm{UL}}_Y) = \mathrm{Supp}(P^{S_\mathrm{L}}_Y)$.

\noindent \textbf{Domain correlated} 
A more challenging but common setting is that unlabeled and labeled data share the same domain space while there is no overlap between the category space of unlabeled and labeled data.
Formally, $\mathrm{Supp}(P^{S_\mathrm{UL}}_D) = \mathrm{Supp}(P^{S_\mathrm{L}}_D)$, $\mathrm{Supp}(P^{S_\mathrm{UL}}_Y) \cap \mathrm{Supp}(P^{S_\mathrm{L}}_Y) = \emptyset$.

\noindent \textbf{Category correlated} Similar with \textit{domain correlated}, this setting assumes that unlabeled and labeled data share the same category space while there is no overlap between the domain space of unlabeled and labeled data.
Formally, $\mathrm{Supp}(P^{S_\mathrm{UL}}_D) \cap \mathrm{Supp}(P^{S_\mathrm{L}}_D) = \emptyset$, $\mathrm{Supp}(P^{S_\mathrm{UL}}_Y) = \mathrm{Supp}(P^{S_\mathrm{L}}_Y)$.

\noindent \textbf{Uncorrelated} When extra data from the same sources (domains) as labeled data are unavailable, there may be no overlap between the category and domain spaces of unlabeled data and labeled data, resulting in the most challenging and flexible setting.
Formally, $\mathrm{Supp}(P^{S_\mathrm{UL}}_D) \cap \mathrm{Supp}(P^{S_\mathrm{L}}_D) = \emptyset$, $\mathrm{Supp}(P^{S_\mathrm{UL}}_Y) \cap \mathrm{Supp}(P^{S_\mathrm{L}}_Y) = \emptyset$.

\subsection{Domain-irrelevant unsupervised learning}\label{para:method}
We propose the Domain-Aware Representation LearnING (DARLING) algorithm for unsupervised domain generalization. 
Generally, we pretrain DARLING on the unlabeled dataset $S_\mathrm{UL}$ before finetuning with the labeled dataset $S_\mathrm{L}$. The finetuning phase can be considered as a standard DG setting thus any DG method such as \cite{huang2020self, matsuura2020domain, qiao2020learning} can be applied. We focus on exploring how unsupervised learning helps models generalize to unseen domains.

Let $S_\mathrm{UL} = \{X_n, d_n\}_{n=1}^N$ be the unlabeled dataset with a set of $N$ images with domain labels but without the ground-truth category labels. \cite{tsai2021mice} has shown that the traditional contrastive learning could be modeled by:

\begin{figure}[t]
    \centering
    \begin{subfigure}[b]{0.101\linewidth}
        \centering
        \includegraphics[width=0.7\linewidth]{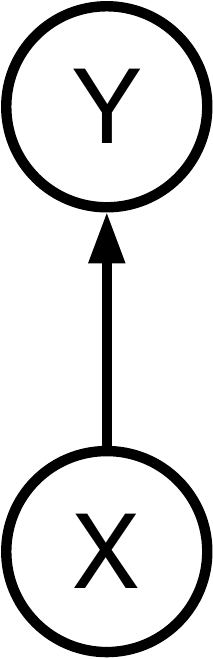}
        \caption{}
        \label{fig:x-y}
    \end{subfigure}
    \,\,\,\,\,\,\,
    \begin{subfigure}[b]{0.3\linewidth}
        \centering
        \includegraphics[width=0.7\linewidth]{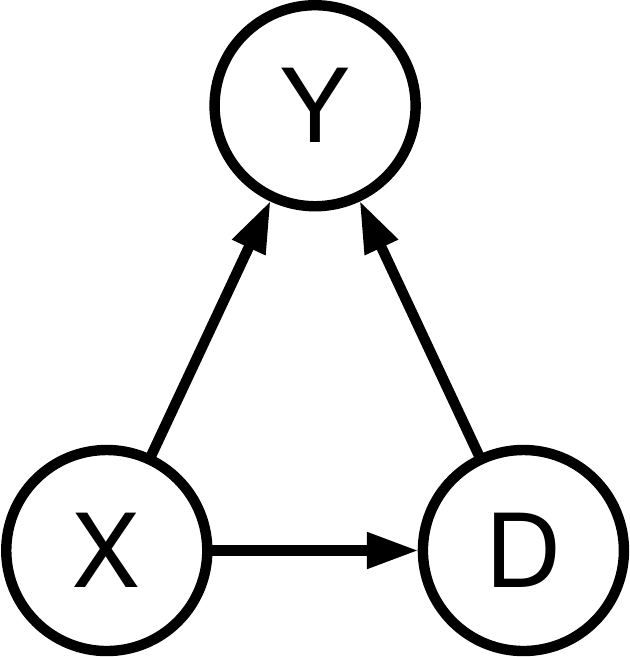}
        \caption{}
        \label{fig:x-y-d}
    \end{subfigure}
    \caption{The graphical model of traditional contrastive learning (a) and our method (b).}
    \label{fig:graph}
    \vspace{-8pt}
\end{figure}

\begin{equation}
\small{
\label{eqn:CL}
    P(\mathbf{Y}|\mathbf{X}) = \prod_{n=1}^N p(y_n | X_n) = \prod_{n=1}^N \frac{\exp (\mathbf{v}_{y_n}^{\top} \mathbf{f}_n / \tau)}{\sum_{i=1}^N \exp(\mathbf{v}_i^{\top} \mathbf{f}_n / \tau)}.
    }
\end{equation}
Here each of the datapoint $X_n$ is assigned with a unique surrogate label $y_n \in \{1, 2, \cdots, N\}$. $\mathbf{v}_{y_n}$ and $\mathbf{f}_n$ are given by passing the input image $\mathbf{X}_n$ to two encoder networks $f_{\mathbf{\theta}}$ and $f_{\mathbf{\theta^{'}}}$. $\tau$ is the temperature hyper-parameter that controls the concentration level. The graphical model is shown in Fig. \ref{fig:x-y}.

\begin{figure*} [th]
    \centering
    \includegraphics[width=0.85\textwidth]{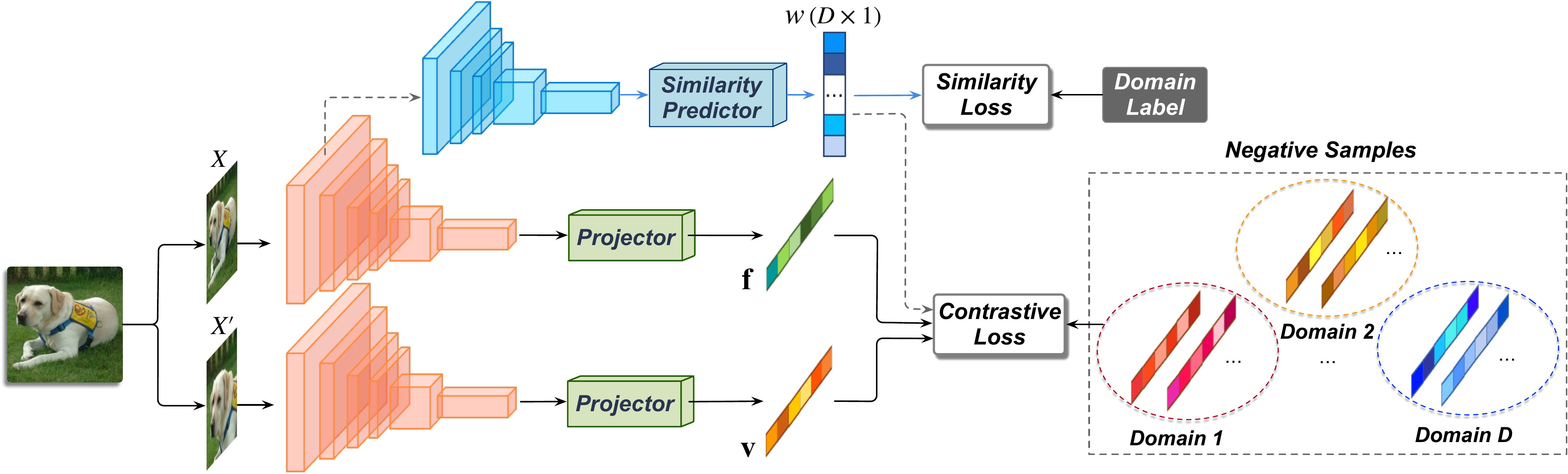}
    \caption{Structure of the proposed DARLING. The upper branch learns domain similarity for a given $X_n$ across all domains and the backpropagation of similarity loss is limited inside the branch (marked with blue arrows). The similarity loss contributes no supervision signal to the training of main network while the contrastive loss does not contribute the learning of similarity branch (connection marked with dashed line).}
    \label{fig:framework}
    \vspace{-8pt}
\end{figure*}

The conditional probability given by Eqn. \ref{eqn:CL} leads to the standard contrastive learning loss. In particular, MoCo learns Eqn. \ref{eqn:CL} via InfoNCE loss by sampling negative samples as follows:
\begin{equation}\label{eq:contrastive loss1}
    \small
    \mathcal{L}(\theta, \theta^{\prime}) = -\frac{1}{N} \sum_{n=1}^{N} \log \frac{\exp \left(\mathbf{v}_{n}^{\top} \mathbf{f}_{n} / \tau\right)}{\exp \left(\mathbf{v}_{n}^{\top} \mathbf{f}_{n} / \tau\right)+\sum_{k=1}^{K} \exp \left(\mathbf{q}_{k}^{\top} \mathbf{f}_{n} / \tau\right)}.
\end{equation}
Here $\mathbf{q} \in \mathbb{R}^{K \times d}$ is a queue of negative samples with size $K$ storing previous embeddings from $f_{\mathbf{\theta^{'}}}$.

However, traditional contrastive learning fails to model domain information. Specifically, the classifier $P(Y|\mathbf{X})$ may be different under different domain label $D$, which leads to model misspecification. Hence given domain label information, we consider the new graphical probability model described in Fig. \ref{fig:x-y-d}.

Next, we give the concrete form of the two conditional probabilities as follows, the generation process of $Y$ is given by: 
\begin{equation} \label{eqn:yxd}
    \small{
    \begin{aligned}
        P(y_n|X_n, D=d) = & \frac{\exp (\mathbf{v}_{y_n, d}^{\top} \mathbf{f}_n / \tau)}{\sum_{i \in N_d} \exp(\mathbf{v}_{i, d}^{\top} \mathbf{f}_n / \tau)} \\
        = & \frac{\exp (\mathbf{v}_{y_n}^{\top} \mathbf{f}_n / \tau)}{\sum_{i \in N_d} \exp(\mathbf{v}_{i}^{\top} \mathbf{f}_n / \tau)},
    \end{aligned}
    }
\end{equation}
where $N_d$ is a collection of training sample indices which belongs to domain $d$. The second equation holds as we further assume the dictionary vectors $\mathbf{v}_{i, d}$ are domain-irrelevant, \textit{i.e.} $\mathbf{v}_{i, d} = \mathbf{v}_{i}$, which could be modeled by a single neural network across all domains.

And the generation process of $D$ is given by:
\begin{equation}
\small
\label{eqn:dx}
    P(D=d | X_n) = \mathrm{softmax}(h(X_n; \Phi))_d,
\end{equation}
where $h$ could be represented as a learnable convolutional neural network parameterized by $\Phi$. Specifically, as shown in Fig. \ref{fig:framework}, features output of shallow layers of the encoder are fed into a stack of extra convolutional networks and a similarity predictor to learn domain similarity for current input. We adopt cross-entropy as the similarity loss.

Hence the likelihood could be obtained from combining Eqn. \ref{eqn:yxd} and Eqn. \ref{eqn:dx} as follows:
\begin{equation} \label{eqn:likelihood}
    {
    \small{
    \begin{aligned}
        P(y_n|X_n) &= \mathbb{E}_{D \sim P(D|X_n)} P(y_n|X_n, D) \\
        &= \sum_d P(D=d | X_n) P(y_n | X_n, D=d) \\
        &= \sum_d w_{n,d} \frac{\exp (\mathbf{v}_{y_n}^{\top} \mathbf{f}_n / \tau)}{\sum_{i \in N_d} \exp(\mathbf{v}_{i}^{\top} \mathbf{f}_n / \tau)}.
    \end{aligned}
    }
    }
\end{equation}
Here $w_{n,d} = P(D=d | X_n)$ is given by Eqn. \ref{eqn:dx}. Noticing that $w_{n, d}$ implies the similarity between domains via each sample, hence Eqn. \ref{eqn:likelihood} eliminates the relevance of domains by reweighting loss on different domains.

We maintain $K$ negative samples $\mathbf{q}_1, \mathbf{q}_2, \dots, \mathbf{q}_K$ and split them into $D$ parts $\mathcal{Q}_1, \mathcal{Q}_2, \dots, \mathcal{Q}_D$ \textit{w.r.t.} their domains.
To be specific, let $e_k \in [D]$ represent the domain of the negative sample $\mathbf{q}_k$. Then $\mathcal{Q}_d$ ($\forall d \in [D]$) can be written as $\{\mathbf{q_k} | e_k = d\}$.
As a result, similar to Eqn. \ref{eq:contrastive loss1}, we write our loss function as:
\begin{equation}\label{eq:contrastive loss2}
    \footnotesize
    \begin{aligned}
        & \mathcal{L}(\theta, \theta^{\prime}) \\
        = & -\frac{1}{N} \sum_{n=1}^{N} \log \sum_d \frac{w_{n, d} \cdot \exp \left(\mathbf{v}_{n}^{\top} \mathbf{f}_{n} / \tau\right)}{\exp \left(\mathbf{v}_{n}^{\top} \mathbf{f}_{n} / \tau\right)+\sum_{\mathbf{q} \in \mathcal{Q}_d} \exp \left(\mathbf{q}^{\top} \mathbf{f}_{n} / \tau\right)}.
    \end{aligned}
\end{equation}
In the processing of optimization, we first learn $\Phi$ by Eqn.\ref{eqn:dx}. With a given $\Phi$ we optimize $\theta$ via minimizing Eqn. \ref{eq:contrastive loss2} until convergence.

\subsection{Domain Specific Negative Samples}

\begin{table*}[ht]
    \centering
    \caption{Results of the \textsl{all correlated} setting on DomainNet. We reimplement state-of-the-art unsupervised methods on DomainNet with ResNet18 \cite{he2016deep} as the backbone network for all the methods unless otherwise specified. ERM indicates the randomly initialed ResNet18. Overall and Avg. indicate the overall accuracy of all the test data and the arithmetic mean of the accuracy of 3 domains, respectively. Note that they are different because the capacities of different domains are not equal. The reported results are average over three repetitions of each run. All the models are trained on `Painting', `Real', and `Sketch' domains of DomainNet and tested on the other three domains. The title of each column indicates the name of the domain used as target. All the models are pretrained for 1000 epoches before finetuned on the labeled data. The best results of all methods are highlighted with the bold font.}
    \resizebox{0.8\textwidth}{!}{
    \begin{tabular}{c|ccc|cc|ccc|cc}
        \toprule
         & \multicolumn{5}{c|}{Label Fraction 1\%} & \multicolumn{5}{c}{Label Fraction 5\%} \\
         \midrule
         method & Clipart & Infograph & Quickdraw & Overall & Avg. & Clipart & Infograph & Quickdraw & Overall & Avg. \\
        \midrule
    
        MoCo V2 \cite{chen2020improved, he2020momentum}  & 18.85 & 10.57 & 6.32 & 10.05  & 11.92 & 28.13 & 13.79 & 9.67 & 14.56 & 17.20 \\
        SimCLR V2 \cite{chen2020big}  & \textbf{23.51} & \textbf{15.42} & 5.29 & 11.80 & \textbf{14.74} & 34.03 & 17.17 & \textbf{10.88} & 17.32 & 20.69 \\
        BYOL \cite{grill2020bootstrap} & 6.21 & 3.48 & 4.27 & 4.45 & 4.65 & 9.60 & 5.09 & 6.02 & 6.49 & 6.90 \\
        AdCo \cite{hu2020adco}  & 16.16 & 12.26 & 5.65 & 9.57 & 11.36 & 30.77 & 18.65 & 7.75 & 15.44 & 19.06 \\
        ERM  & 6.54 & 2.96 & 5.00 & 4.75 & 4.83 & 10.21 & 7.08 & 5.34 & 6.81 & 7.54 \\
        \midrule
        DARLING (ours)  & 18.53 & 10.62 & \textbf{12.65} & \textbf{13.29} & 13.93 & \textbf{39.32} & \textbf{19.09} & 10.50 & \textbf{18.73} & \textbf{22.97} \\
        
        \midrule
        & \multicolumn{5}{c|}{Label Fraction 10\%} & \multicolumn{5}{c}{Label Fraction 100\%} \\
         \midrule
         method & Clipart & Infograph & Quickdraw & Overall & Avg. & Clipart & Infograph & Quickdraw & Overall & Avg. \\
        \midrule

        MoCo V2 & 32.46 & 18.54 & 8.05 & 15.92 & 19.69 & 64.18 & 27.44 & 25.26 & 33.76 & 38.96 \\
        SimCLR V2 & \textbf{37.11} & 19.87 & 12.33 & 19.45 & 23.10 & 68.72 & 27.60 & 30.56 & 37.47 & 42.29 \\
        BYOL & 14.55 & 8.71 & 5.95 & 8.46 & 9.74 & 54.44 & 23.70 & 20.42 & 28.23 & 32.86 \\
        AdCo & 32.25 & 17.96 & 11.56 & 17.53 & 20.59 & 62.84 & 26.69 & 26.26 & 33.80 & 38.60 \\ 
        ERM  & 15.10 & 9.39 & 7.11 & 9.36 & 10.53 & 52.79 & 23.72 & 19.05 & 27.19 & 31.85 \\
        \midrule
        DARLING (ours)  & 35.15 & \textbf{20.88} & \textbf{15.69} & \textbf{21.08} & \textbf{23.91} & \textbf{72.79} & \textbf{32.01} & \textbf{33.75} & \textbf{41.19} & \textbf{46.18} \\
        \bottomrule
    \end{tabular}}
    \label{tab:all_correlated_domainet}
    \vspace{-0.28cm}
\end{table*}

As shown in Eqn. \ref{eq:contrastive loss2}, we maintain domain specific negative queries to calculate similarity across domains.
Inspired by \cite{hu2020adco}, we propose a domain specific negative samples generating mechanism with a adversarial updating manner to closely track the change of representations for each domain. 
Our objective can be considered as:

\begin{equation}
    \small
    \theta^{\star}, \mathcal{Q}_1^{\star}, ..., \mathcal{Q}_D^{\star}=\arg \min _{\theta} \max _{\mathcal{Q}_1, \dots, \mathcal{Q}_D} \mathcal{L}(\theta, \mathcal{Q}_1, ..., \mathcal{Q}_D).
\end{equation}
Specifically, we iteratively update network weights $\theta$ and domain-specific negative adversaries as follows:

\begin{equation}
{\small
    \left\{
    \begin{aligned}
        \theta & \longleftarrow \theta-\eta_{\theta} \frac{\partial \mathcal{L}(\theta, \mathcal{Q}_1, ..., \mathcal{Q}_D)}{\partial \theta}, \\
        \mathbf{q}_{k} & \longleftarrow \mathbf{q}_{k}+\eta_{\mathcal{N}} \frac{\partial \mathcal{L}(\theta, \mathcal{Q}_1, ..., \mathcal{Q}_D)}{\partial \mathbf{q}_{k}},
    \end{aligned}
    \right.
}
\end{equation}
where $k = 1, 2, \dots, K$ is the index of negative samples, $\eta_{\theta}$ and $\eta_{\mathcal{N}}$ are the learning rates for network weights and negative adversaries, respectively.

To simplify the calculation of negative samples, here we constrict $w_{n,d} = \mathbb{I}[d_n = d]$, thus a given sample only contributes to the generation of negative samples from the same domain. Under this circumstance, the loss function in Eqn. \ref{eq:contrastive loss2} could be written as
\begin{equation}
    \footnotesize
    \begin{aligned}
        & L(\theta, \theta') \\
        = & -\frac{1}{N}\sum_{n=1}^N\log \frac{\exp \left(\mathbf{v}_{n}^{\top} \mathbf{f}_{n} / \tau\right)}{\exp \left(\mathbf{v}_{n}^{\top} \mathbf{f}_{n} / \tau\right)+\sum_{\mathbf{q} \in \mathcal{Q}_{d_n}} \exp \left(\mathbf{q}^{\top} \mathbf{f}_{n} / \tau\right)}.
    \end{aligned}
\end{equation}
And the derivative of $\mathcal{L}$ in updating negative sample $\mathbf{q}_k$ is
\begin{equation}
    \small
    \frac{\partial \mathcal{L}}{\partial \mathbf{q}_k} = \frac{1}{N\tau}\sum_{n=1}^N s_{nk} \cdot \mathbf{f}_n,
\end{equation}
where
\begin{equation}
    \footnotesize
    s_{nk}= \begin{dcases}
        \frac{\exp \left(\mathbf{q}_{k}^{\top} \mathbf{f}_{n} / \tau\right)}{\exp \left(\mathbf{v}_{n}^{\top} \mathbf{f}_{n} / \tau\right)+\sum_{\mathbf{q} \in \mathcal{Q}_{d_n}} \exp \left(\mathbf{q}^{\top} \mathbf{f}_{n} / \tau\right)}, & \text{if } e_k = d_n, \\
    0, & \text{otherwise.}
    \end{dcases}
\end{equation}

Our objective is to maintain hard samples for positive ones in each domain, so the negative samples in a given domain are pushed closer towards the queries from the same domain, thus ${n}_{k,d}$ is optimized to maximize the similarities between them and positive queries within the corresponding domains.

The superiority of the proposed domain specific negative sample generation is two fold. Firstly, it yields constant number of negative samples while other updating methods such as \cite{chen2020simple} may yield various number of samples from different domains. Secondly, the proposed method generates hard negative samples with the most confusing negative pairs within each domain, which is consistent with Eqn. \ref{eq:contrastive loss2}.

\section{Experiments} \label{exp}
\begin{table*}[ht]
    \centering
    \caption{Results of the \textsl{all correlated} setting on PACS. Given the experiment for each target domain is run respectively, there is no overall accuracy across domains. Thus we report the average accuracy and the accuracy for each domain. For details about the number of runs, meaning of column titles and fonts, see Table \ref{tab:all_correlated_domainet}.}
    \resizebox{0.7\textwidth}{!}{
    \begin{tabular}{c|cccc|c|cccc|c}
        \toprule
         & \multicolumn{5}{c|}{Label Fraction 1\%} & \multicolumn{5}{c}{Label Fraction 5\%} \\
         \midrule
         method & Photo & Art. & Cartoon & Sketch & Avg. & Photo & Art. & Cartoon & Sketch & Avg. \\
        \midrule
        MoCo V2 & 22.97 & 15.58 & 23.65 & 25.27 & 21.87 & 37.39 & 25.57 & 28.11 & 31.16 & 30.56 \\
        SimCLR V2 & \textbf{30.94} & 17.43 & \textbf{30.16} & 25.20 & 25.93 & \textbf{54.67} & 35.92 & 35.31 & \textbf{36.84} & \textbf{40.68} \\
        BYOL & 11.20 & 14.53 & 16.21 & 10.01 & 12.99 & 26.55 & 17.79 & 21.87 & 19.65 & 21.47 \\
        AdCo & 26.13 & 17.11 & 22.96 & 23.37 & 22.39 & 37.65 & 28.21 & 28.52 & 30.35 & 31.18 \\ 
        ResNet-18  & 10.90 & 11.21 & 14.33 & 18.83 & 13.82 & 14.15 & 18.67 & 13.37 & 18.34 & 16.13 \\
        \midrule
        DARLING (ours)  & 27.78 & \textbf{19.82} & 27.51 & \textbf{29.54} & \textbf{26.16} & 44.61 & \textbf{39.25} & \textbf{36.41} & 36.53 & 39.20 \\
        \midrule
        & \multicolumn{5}{c|}{Label Fraction 10\%} & \multicolumn{5}{c}{Label Fraction 100\%} \\
         \midrule
         method & Photo & Art. & Cartoon & Sketch & Avg. & Photo & Art. & Cartoon & Sketch & Avg. \\
        \midrule
        MoCo V2 & 44.19 & 25.85 & 33.53 & 24.97 & 32.14 & 59.86 & 28.58 & 48.89 & 34.79 & 43.03 \\
        SimCLR V2 & \textbf{54.65} & 37.65 & 46.00 & 28.25 & 41.64 & 67.45 & \textbf{43.60} & 54.48 & 34.73 & 50.06\\
        BYOL & 27.01 & 25.94 & 20.98 & 19.69 & 23.40 & 41.42 & 23.73 & 30.02 & 18.78 & 28.49 \\
        AdCo & 46.51 & 30.21 & 31.45 & 22.96 & 32.78 & 58.59 & 29.81 & 50.19 & 30.45 & 42.26 \\  
        ResNet-18  & 16.27 & 16.62 & 18.40 & 12.01 & 15.82 & 43.29 & 24.27 & 32.62 & 20.84 & 30.26 \\
        \midrule
        DARLING (ours)  & 53.37 & \textbf{39.91} & \textbf{46.41} & \textbf{30.17} & \textbf{42.47} & \textbf{68.66} & 41.53 & \textbf{56.89} & \textbf{37.51} & \textbf{51.15} \\
        
        \bottomrule
    \end{tabular}
    }
    \label{tab:all_correlated_pacs}
    % \vspace{-10px}
\end{table*}

In this section, we specifically describe experimental settings that support unsupervised domain generalization (UDG) and show experimental results of the proposed DARLING and its state-of-the-art counterparts.

\subsection{Unsupervised Domain Generalization (UDG)}
\paragraph{Settings and datasets} 
We present extensive experimental results on 3 of 4 settings that are more common in real-world scenarios, namely \textsl{all correlated}, \textsl{domain correlated}, and \textsl{uncorrelated}. The correlations between unlabeled and labeled data gradually decrease in these settings. Experiments on the remaining \textsl{category correlated} setting are in Appendix B.1. We adopt four datasets to carry through evaluations, namely DomainNet \cite{peng2019moment}, PACS \cite{li2017deeper}, CIFAR-10-C and CIFAR-100-C \cite{hendrycks2018benchmarking, krizhevsky2009learning}. Introduction to these datasets and details of implementation are in Appendix B.1.

\paragraph{All correlated UDG} We explore how unsupervised learning enhances the generalization ability of models when training data are partially labeled and both the category and domain between unlabeled and labeled data are correlated.  
We adopt DomainNet and PACS for this setting. For DomainNet, we randomly select 3 out of 6 domains as source domains and the remaining as target domains. 20 out of 300 categories are randomly selected for both training and testing. 
For PACS, we follow the common DG setting where one domain is considered as the target domain while the others as source domains for each run. 
The proportion of labeled data to training data for both datasets varies from 1\% to 10\%.

Results are shown in Table \ref{tab:all_correlated_domainet} (DomainNet) and \ref{tab:all_correlated_pacs} (PACS). DARLING outperforms other counterparts with all given fractions of labeled data on average accuracy on both DomainNet and PACS. Surprisingly, when all the training data are labeled, unsupervised pretraining with the same data improves the prediction accuracy on target domains by a noticeable margin. This indicates that when there are severe distribution shifts between training and testing data, the supervision from category labels of source domains is insufficient given that it can be considered as biased knowledge in target domains. Unsupervised learned dissimilarities among samples from the same category in source domains can introduce valid knowledge for distinguishing categories in target domains, for which unsupervised learning naturally fits the DG problem.

Moreover, from the perspective of the graphical model mentioned in Section \ref{para:method}, supervision from the source domain helps the model to learn a domain-relevant classifier, which can fail in target domains. While DARLING learns a domain-irrelevant representation space leading to more robust predictions in novel domains. Thus DARLING achieves a significant improvement compared to SOTA unsupervised learning methods (7.43\% compared to MoCo V2 and 3.89\% to SimCLR V2).
When the fraction of labeled data is lower than 10\%, we only finetune the linear classifier for all the methods to prevent overfitting.
Both random initialized ResNet-18 and BYOL fail to learn a valid model with label fractions of 1\% and 5\%, while DARLING consistently achieves considerable improvement. 
Here we report the results of one of the possible divisions. Details of the data partition and results of other divisions are in Appendix B.2.

\begin{table*}[ht]
    \centering
    \caption{Results of the \textsl{domain correlated} setting on DomainNet. For details about meaning of column titles and fonts, see Table \ref{tab:all_correlated_domainet}.}
    \resizebox{0.80\textwidth}{!}{
    \begin{tabular}{c|ccc|cc|ccc|cc}
        \toprule
         & \multicolumn{5}{c|}{Pretraining with data from 40 categories} & \multicolumn{5}{c}{Pretraining with data from 100 categories} \\
         \midrule
         method & Clipart & Infograph & Quickdraw & Overall & Avg. & Clipart & Infograph & Quickdraw & Overall & Avg. \\
        \midrule
        MoCo V2 & 72.84 & 33.40 & 34.20 & 41.19 & 46.82 & 77.03 & 37.68 & 35.25 & 43.71 & 49.98 \\
        SimCLR V2 & 75.58 & \textbf{35.52} & 37.08 & 43.83 & 49.39 & 79.70 & 38.88 & 38.89 & 46.50 & 52.49 \\
        BYOL & 58.39 & 23.99 & 28.56 & 32.87 & 36.98 & 58.27 & 24.14 & 27.83 & 32.49 & 36.75 \\
        AdCo & 76.61 & 31.55 & 33.42 & 40.96 & 47.19 & 75.19 & 33.76 & 38.51 & 43.77 & 49.15 \\ 
        ERM & 55.78 & 22.40 & 25.75 & 30.43 & 34.64 & 55.78 & 22.40 & 25.75 & 30.43 & 34.64 \\
        \midrule
        DARLING (ours)  & \textbf{78.40} & 33.98 & \textbf{39.87} & \textbf{44.20} & \textbf{50.75} & \textbf{82.28} & \textbf{40.60} & \textbf{47.68} & \textbf{52.19} & \textbf{56.85} \\
        \bottomrule
    \end{tabular}}
    \label{tab:domain_correlated_domainnet}
    % \vspace{-10px}
\end{table*}

\begin{table}[ht]
    \centering
    \caption{Results of the \textsl{uncorrelated} setting on CIFAR. Pre. and Fine. are short for pretraining and finetuning data. All methods are pretrained on domain `elastic', `fog', `impulse noise' and `motion blur', and fine-tuned on domain `contrast', `frost', `glass', `blur' and `shot noise'. 
    For details about the number of runs, meaning of column titles and fonts, see Table \ref{tab:all_correlated_domainet}.}
    \resizebox{1\linewidth}{!}{
    \begin{tabular}{c|cccc|c}
        \toprule
         method & Brightness & Defocus Blur & Gaussian Noise & Snow & Avg. \\
        \midrule
        MoCo V2 & 77.13 & 75.88 & 75.18 & 72.27 & 75.12 \\
        SimCLR V2 & 78.54 & 77.60 & 75.92 & 72.68 & 76.19 \\
        BYOL & 58.10 & 57.07 & 56.31 & 53.96 & 56.36 \\
        AdCo & 75.63 & 77.32 & 78.84 & 72.25 & 76.01 \\
        ERM & 36.53 & 34.61 & 33.49 & 32.98 & 34.40 \\
        \midrule
        DARLING (ours) & \textbf{80.28} & \textbf{77.74} & \textbf{79.65} & \textbf{77.76} & \textbf{78.86} \\
        \bottomrule
    \end{tabular}
    }
    \label{tab:uncorrelated_cifar}
\end{table}

\begin{table}[t]
    \centering
    \caption{Results of state-of-the-art methods with different initialization methods under the \textsl{domain correlated} setting on DomainNet. For details about the number of runs, meaning of column titles and fonts, see Table \ref{tab:all_correlated_domainet}.}
    \resizebox{\linewidth}{!}{
    \begin{tabular}{c|ccc|cc}
        \toprule
        method & Clipart & Infograph & Quickdraw & Overall & Avg. \\
        \midrule
        M-ADA \cite{qiao2020learning} & 65.33 & 37.51 & 30.16 & 38.75 & 44.33 \\
        RSC \cite{huang2020self} & 61.25 & 23.27 & 27.48 & 31.52 & 37.33 \\
        MMLD \cite{matsuura2020domain}  & 74.09 & 36.09 & 33.44 & 42.46 & 47.88 \\
        MoCo V2 + RSC   & 81.36 & 37.59 & 41.38 & 46.81 & 53.44 \\
        MoCo V2 + MMLD  & 82.46 & \textbf{39.52} & 40.58 & 47.83 & 54.19 \\ 
        \midrule
        DARLING (ours) + RSC & \textbf{86.47} & 39.00 & \textbf{45.71} & \textbf{49.71} & \textbf{57.06} \\
        DARLING (ours) + MMLD & 85.53 & 38.14 & 44.08 & 48.62 & 55.92 \\
        \bottomrule
    \end{tabular}
    }
    \label{tab:dg_domainnet}
    \vspace{-10px}
\end{table}

\paragraph{Domain correlated UDG}
\textsl{Domain correlated UDG} is a challenging setting with a great degree of flexibility, where unlabeled data can be sampled from other categories or even other datasets compared with labeled data as long as they share the same domain space. This setting is quite common in real-world scenarios, given that when category space is unknown, one can hardly assume that the unlabeled data share the same categories with labeled data. 
We use this setting to validate the generalization ability of unsupervised learning methods under both domain and category shifts.

We adopt DomainNet for this setting, given that the category spaces of other popular DG datasets (such as PACS, VLCS \cite{torralba2011unbiased} and Office-home \cite{venkateswara2017deep}) are limited. We randomly select 3 out of 6 domains as source domains, and the remaining domains are considered as targets. We choose 20 out of 300 categories for labeled training and testing data and the other 40 or 100 categories for unlabeled data. There is no overlap between categories in unlabeled and labeled data, leading to the most challenging scenario in this setting. Details of data proportion and more experimental results are in Appendix B.3.

Results are shown in Table \ref{tab:domain_correlated_domainnet}. DARLING achieves the highest generalization accuracy on all of the domains. As aforementioned, current contrastive loss not only enlarges the distance between representations of samples from different categories but also that of samples from different domains. However, the representations of domains being more distinguishable brings no benefit on downstream tasks \cite{muandet2013domain}. On the contrary, DARLING forces the model to learn a domain-irrelevant representation space where only representations from different latent categories can be easily distinguished. Intuitively, DARLING learns two kinds of abilities: 1) selecting domain-irrelevant features which are most likely related to categories, and generating a latent representation space with them; 2) discerning domain-related features and preventing them from contributing to the representation space. As a result, DARLING shows strong generalization ability under both domain and category shifts.

\paragraph{Uncorrelated UDG} In this setting, we make no restrictions or assumptions about the correlation between unlabeled and labeled data. Thus unlabeled data can be sampled from novel domains, unknown categories or other datasets compared with labeled data. This brings a great challenge to the generalization ability of models and the effectiveness of unsupervised learning, given that the mutual information between unlabeled and labeled data can reach the minimum. 
Intuitively, with a stronger connection between unlabeled and labeled data, unsupervised learning on unlabeled data brings greater improvement. We explore how unsupervised learning affects the generalization ability of models to novel domains when the distribution shifts between unlabeled and labeled data are significant.

Since the domain spaces of DomainNet and PACS are limited, we adopt CIFAR-100-C and CIFAR-10-C for this setting. In the most challenging scenario, the distribution of unlabeled data and labeled data can be uncorrelated, where we consider unlabeled CIFAR-100-C as the pretraining data and CIFAR-10-C for finetuning data and target data. To make the domain space sufficient, we generate distinguishing domains for CIFAR-100 and CIFAR-10 and, there is no overlap among unlabeled training data, labeled training data, and test data. 
As shown in Table \ref{tab:uncorrelated_cifar}, we randomly select 4 specific domains for pretraining, finetuning, and test data, respectively, and set the severity level to 3. We adopt ResNet18 for this setting and the first layer was replaced by a convolution layer with a kernel size of 3 and stride of 1, since the size of images from CIFAR is $32 \times 32$. Details of implementation and further experimental results when domains are differently selected are in Appendix B.4.

Results are shown in Table \ref{tab:uncorrelated_cifar}. Surprisingly, unsupervised pretraining methods achieve significant improvement though the pretraining data provides limited knowledge about labeled training and testing data.
DARLING outperforms all the unsupervised learning counterparts on all the domains by a noticeable margin. The superiority of DARLING shows that samples even from different domains and categories compared to test data can help the model distinguish domain correlated features and category-correlated features in the finetuning phase. In other words, similarities among category correlated features may help select predictive features, while similarities among domain correlated features help the models ignore category irrelevant features.  

This largely broadens the valid scope of unsupervised learning, given that no constraints of the category and source (domain) of pretraining data and labeled data are required to improve the model performance on novel domains.

\begin{figure*}[t]
  \centering
  \begin{subfigure}[b]{0.35\textwidth}
        \centering
        \includegraphics[width=\textwidth]{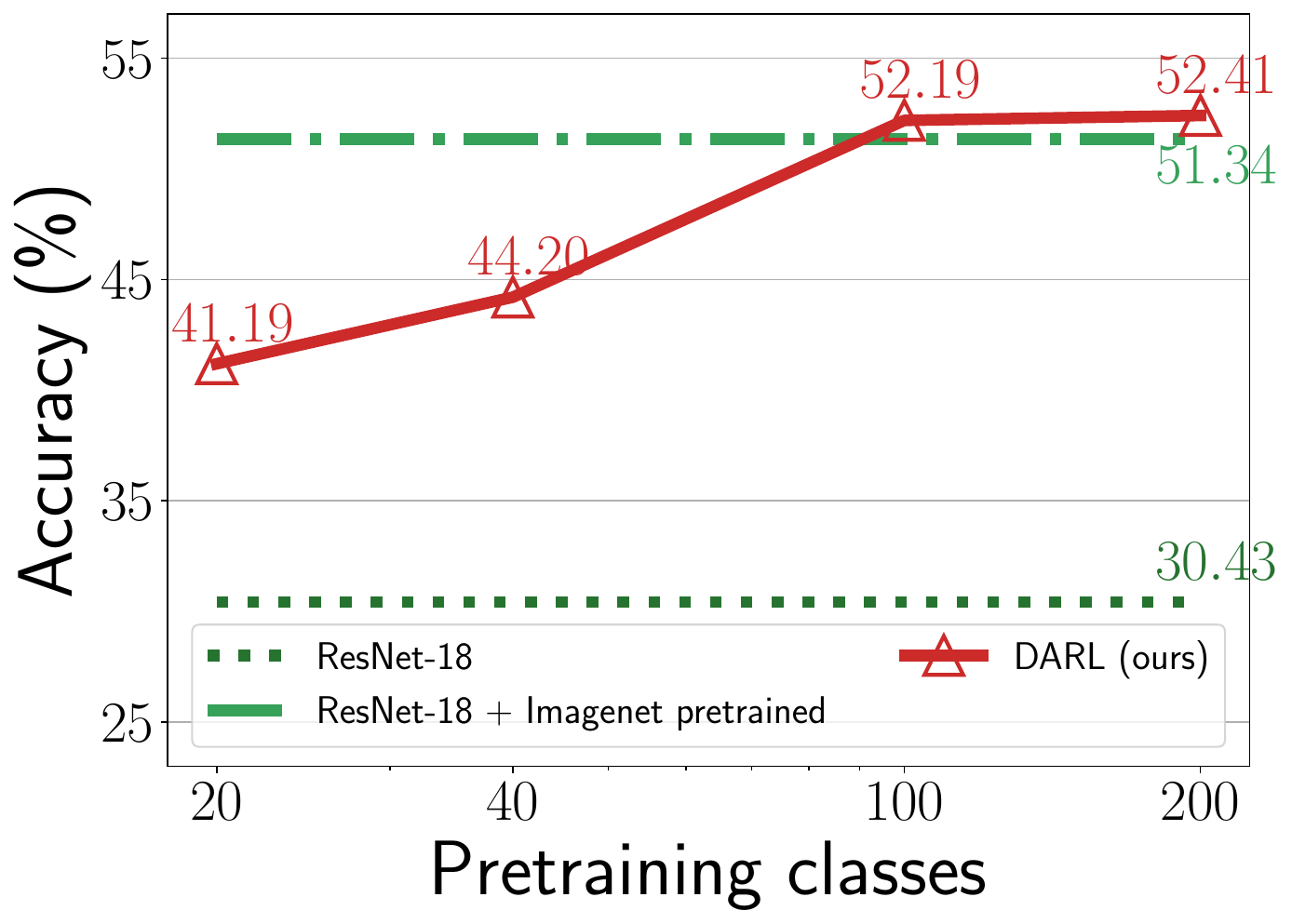}
        \caption{}
        \label{fig:pretrain_classes}
    \end{subfigure}
    \,\,\,\,\,\,\,\,\,\,\,\,\,
    % \hfill
    \begin{subfigure}[b]{0.35\textwidth}
        \centering
        \includegraphics[width=\textwidth]{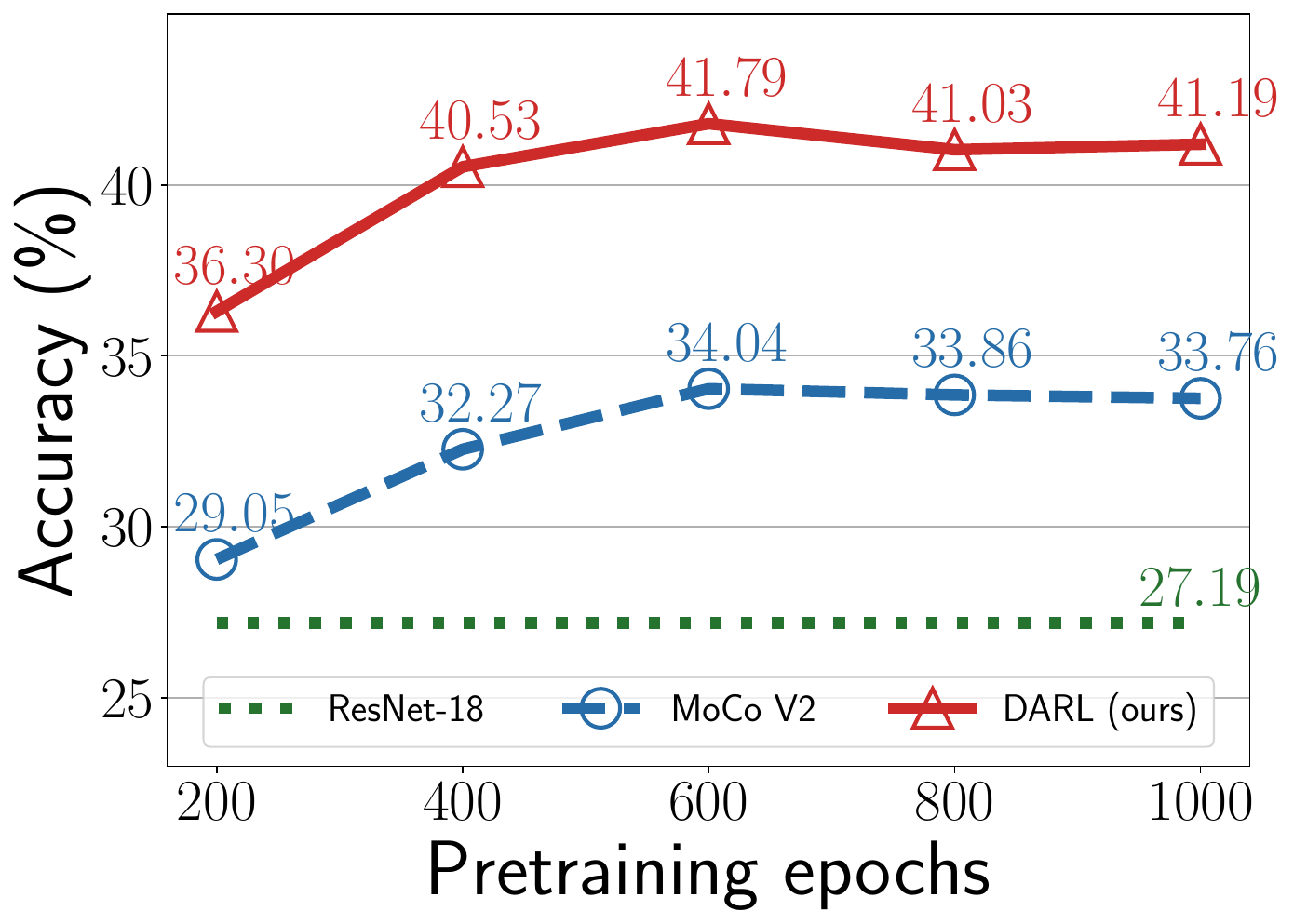}
        \caption{}
        \label{fig:epoch}
    \end{subfigure}
    \hfill
    % \vspace{-10px}
    \caption{Figure (a) shows how the generalization accuracy of DARLING grows as the amount of available data increases. All the models are pretrained with DARLING for 600 epochs. Pretraining classes indicates the number of categories used in the pretraining phase. DARLING outperfroms models with ImageNet pretrained weights as the initialization when the number of available categories reaches 100. Figure (b) shows the convergence speed of MoCo V2 and DARLING. Pretraining epoch is the number of epoch for pretraining. DARLING is more time efficient and achieves considerable higher generalization accuracy after convergence.}
    \label{fig:three graphs}
    % \vspace{-10px}
\end{figure*}

\paragraph{Finetuning with DG methods} Table \ref{tab:dg_domainnet} shows how unsupervised pretraining methods benefit the generalization ability of ERM models since all the finetuning phases of these methods can be considered as the training phase of ERM models. Here we further explore how unsupervised training affects the models trained with effective domain generalization methods. We report the results of state-of-the-art methods of domain generalization with unsupervised trained parameters as the initial state in Table \ref{tab:dg_domainnet}. 
More experimental results are in Appendix B.5.

\subsection{Comparison with ImageNet Pretrained Models}

As the amount of available unlabeled data grows, unsupervised pretraining achieves better performance. Surprisingly, we find it is possible for DARLING to outperform models pretrained on ImageNet though our unlabeled training data is of a significantly smaller amount compared to ImageNet. 
Actually, if we consider ImageNet as a mixture of data sampled from latent domains, the heterogeneity is limited for learning a generalizable model with a domain-irrelevant representation space \cite{he2021towards}.
Given data with strong heterogeneity (such as a subset of DomainNet), although there are strong distribution shifts between training data and testing data, DARLING can still learn domain-irrelevant representations and strengthen the generalization ability to target domains. As shown in Figure \ref{fig:pretrain_classes}, when the available data for pretraining are more than 100 out of 300 categories from DomainNet, DARLING outperforms ImageNet pretrained initialization by a noticeable margin. Note that the number of data that DARLING uses for pretraining is less than $1/10$ of the number of ImageNet data.
This observation indicates that stronger unsupervised pretrained models can be alternatives to the ones pretrained on ImageNet as the initialization approach for DG tasks.

\subsection{Analysis}

Figure \ref{fig:epoch} visualizes the accuracy on the \textsl{all correlated} setting under various pretraining epochs. All the parameters and pretraining protocols used for both methods are the same for a fair comparison. With a small number of pretraining epochs, DARLING outperforms MoCo V2 by a considerable margin. As the number of pretraining epochs grows, MoCo V2 reaches a maximal overall accuracy of 34.04\% after pretraining for 600 epochs before the curve tends to be flat, while DARLING outperforms MoCo V2 by around 7.43\% at epoch 1000. The curve indicates that DARLING is not only an efficient pretraining method but also with a better convergence point. 
Moreover, Figure \ref{fig:pretrain_classes} and \ref{fig:epoch} indicate that with more unlabeled data sampled from different domains and more training epochs, strong UDG methods can gradually improve models' generalization ability. Thus pretraining models on data from different datasets may further enhance generalization ability. 

\section{Conclusion}
In this paper, we proposed a novel problem called unsupervised domain generalization (UDG), where unlabeled data are used to strengthen the generalization ability of models since labeled data are usually costly or unavailable. We also proposed a Domain-Aware Representation Learning method called DARLING to address the UDG problem. Extensive experiments clearly demonstrated the effectiveness of the proposed DARLING compared with state-of-the-art unsupervised learning counterparts. As a pretraining approach, DARLING outperforms ImageNet pretraining approach with significantly less data, showing an encouraging way to initialize models for the DG problem. 

\section*{Acknowledgement}
% \paragraph{Acknowledgement} 
This work was supported in part by National Key R\&D Program of China (No. 2018AAA0102004, No. 2020AAA0106300), National Natural Science Foundation of China (No. U1936219, 61521002, 61772304), Beijing Academy of Artificial Intelligence (BAAI), and a grant from the Institute for Guo Qiang, Tsinghua University.

\clearpage
{\small
\bibliographystyle{ieee_fullname}
\bibliography{ref}
}

\clearpage

\appendix
\onecolumn
\section{Extra Experimental Results and Details}

\subsection{Datasets and details of implementation}
We adopt 4 datasets to conduct experiments in our 4 settings. We briefly introduce them as follows.

\begin{figure*} [th]
    \begin{subfigure}[b]{0.45\textwidth}
        \centering
        \includegraphics[width=\textwidth]{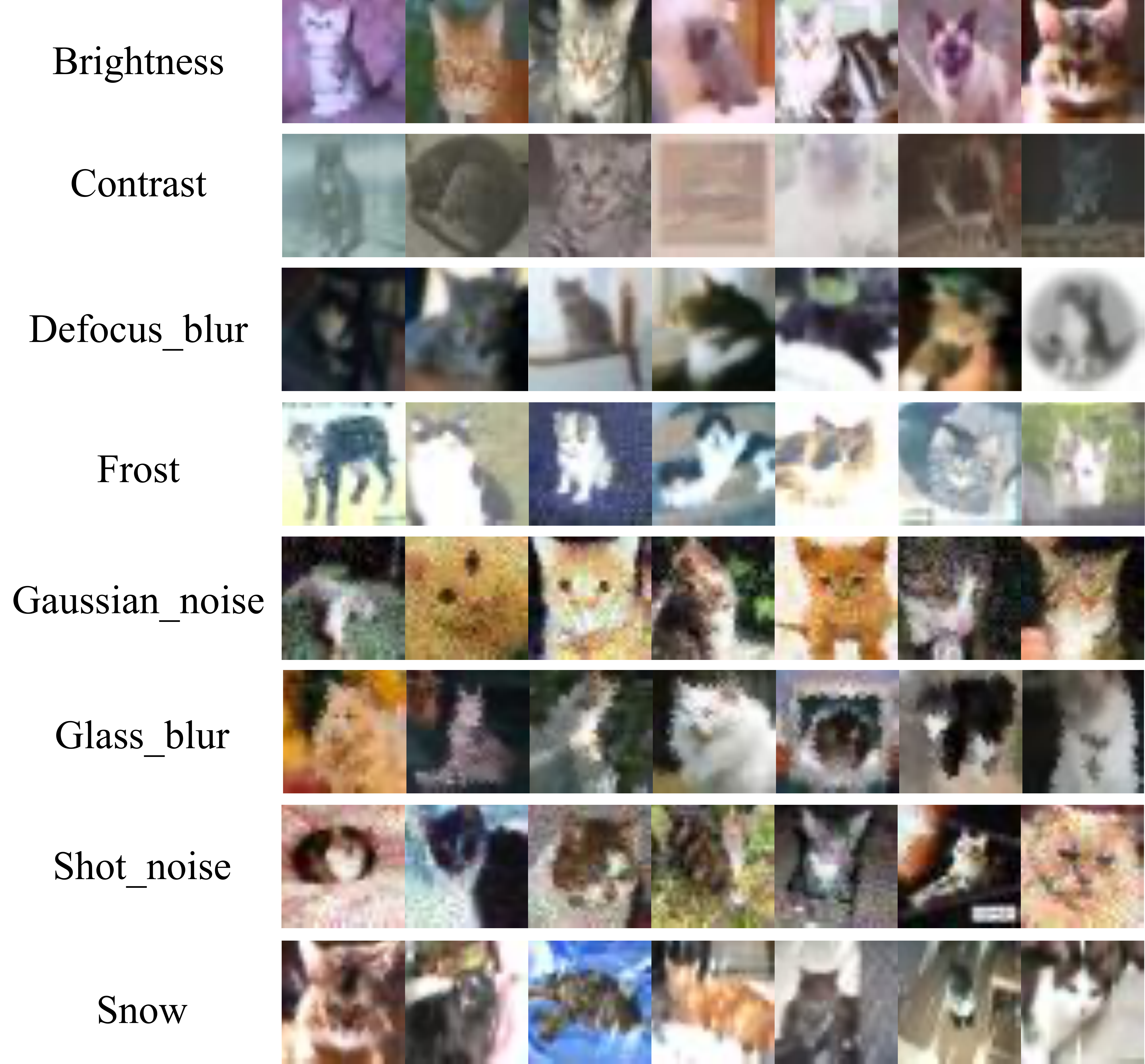}
        \caption{Example images of class cat in Cifar10-C.}
        \label{fig:cifar10}
    \end{subfigure}
    \hfill
    \begin{subfigure}[b]{0.54\textwidth}
        \centering
        \includegraphics[width=\textwidth]{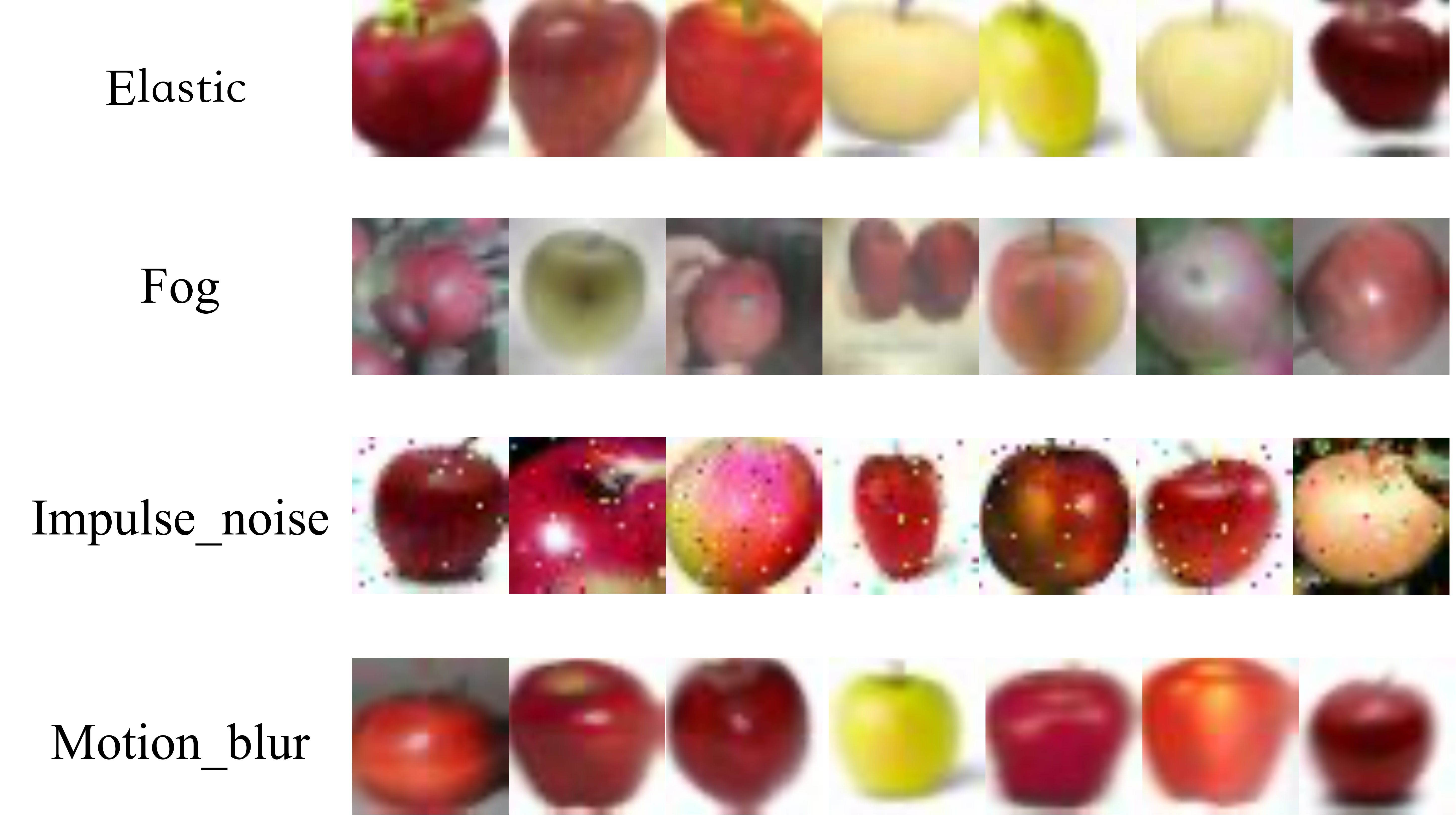}
        \caption{Example images of class apple in Cifar100-C.}
        \label{fig:cifar100}
    \end{subfigure}
    \caption{Example images of Cifar.}
    \label{fig:cifar}
\end{figure*}

\textbf{CIFAR-10-C and CIFAR-100-C} are robustness benchmarks consisting of 19 corruptions types with five levels of severities. We select level 3 for all the experiments. Example images are shown in Fig. \ref{fig:cifar}.

\begin{figure*} [th]
    \centering
    \includegraphics[width=0.8\linewidth]{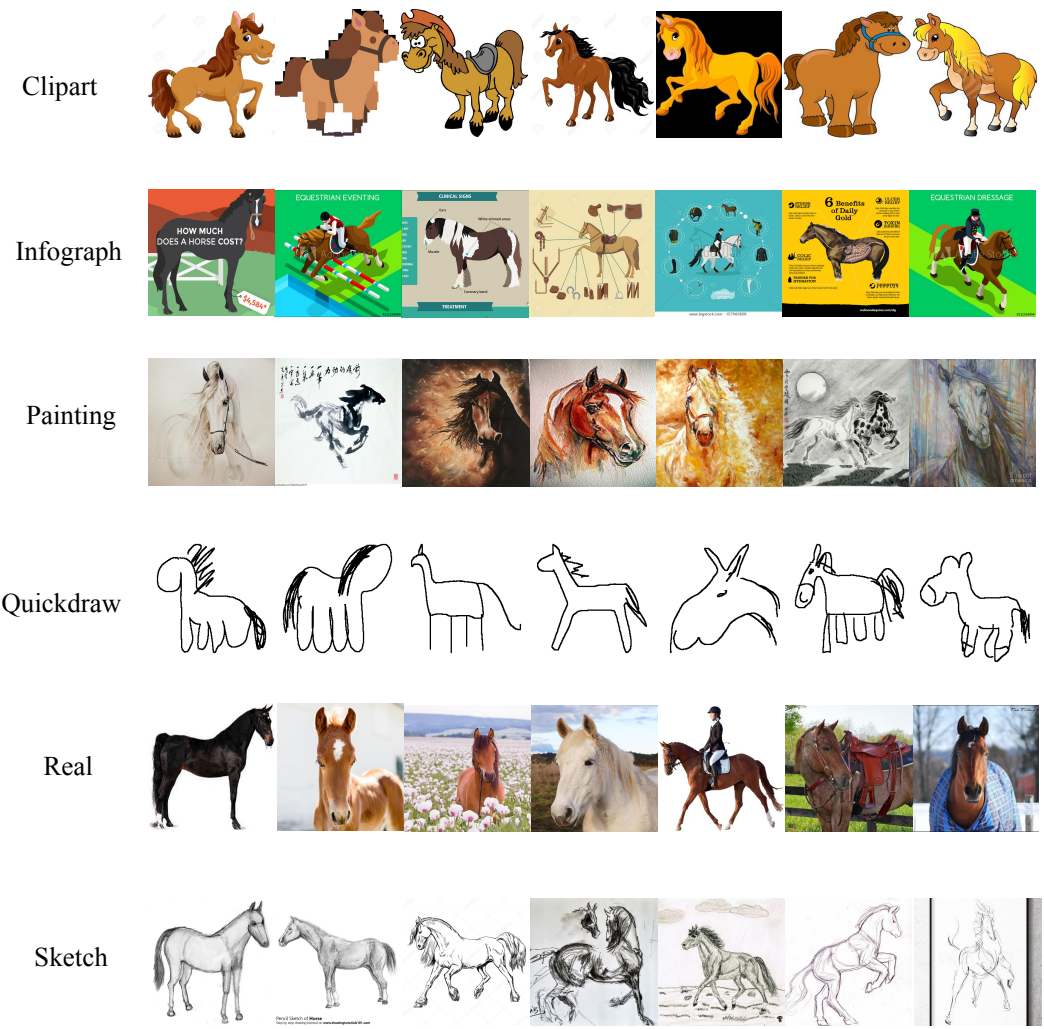}
    \caption{Example images of class horse in DomainNet.}
    \label{fig:domainnet}
\end{figure*}

\textbf{DomainNet} is comprised of 6 domains, namely \textsl{clipart, infograph, painting, quickdraw, real, sketch}. It contains 586, 575 examples of size (3, 224, 224) and 345 classes. Example images are shown in Fig. \ref{fig:domainnet}.

\begin{figure*} [th]
    \centering
    \includegraphics[width=0.8\linewidth]{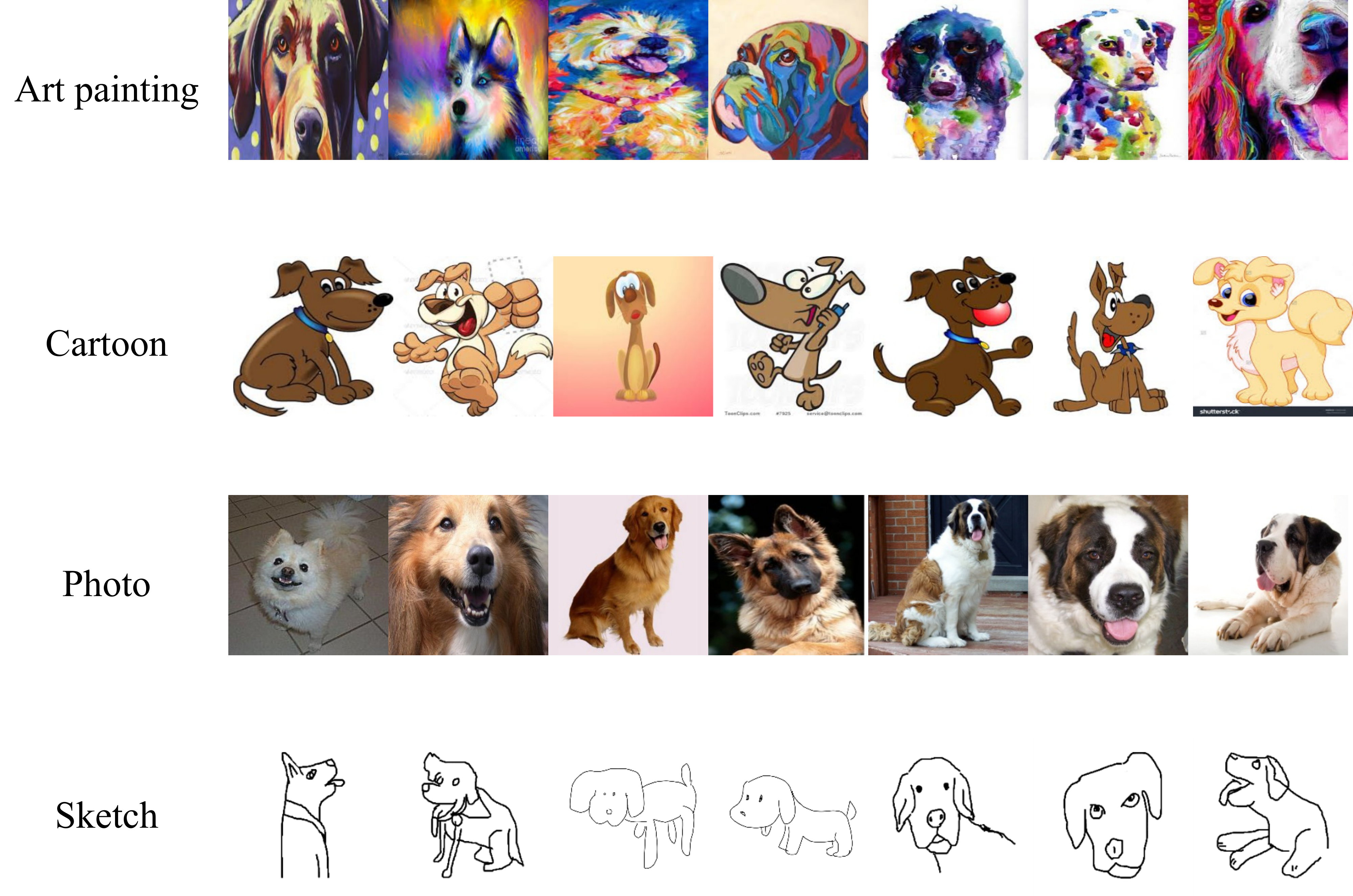}
    \caption{Example images of class dog in PACS.}
    \label{fig:pacs}
\end{figure*}

\textbf{PACS} is a widely used benchmark for domain generalization which consists of 7 object categories spanning 4 image styles, namely \textsl{photo, art-painting, cartoon and sketch}. We adopt the protocol in \cite{li2017deeper} to split the training and val set. Example images are shown in Fig. \ref{fig:pacs}.

\textbf{Details of implementation}. For all the experiments, we use ResNet-18 as the backbone network unless otherwise specified. The learning rate for pretraining is 0.03 and then decays with a cosine decay schedule. The weight decay is set to $1e^{-4}$ and the batch size is set to 1024. For \textsl{domain correlated} and \textsl{category correlated}, all methods are pretrained for 600 epochs, while for other settings 1000 epochs. We follow\cite{chen2020improved} for the augmentations and the optimization temperature. The feature dimension is set to 128. For finetuning, all methods are trained for 30 epochs for all the settings, while the learning rate and weight decay are set to $1e^{-3}$ and $1e^{-4}$, respectively.

\subsection{All correlated UDG}
\begin{table*}[ht]
    \centering
    \caption{Results of the \textsl{all correlated} setting on DomainNet. We reimplement state-of-the-art unsupervised methods on DomainNet with ResNet18 \cite{he2016deep} as the backbone network for all the methods. ERM indicates the randomly initialed ResNet18. Overall and Avg. indicate the overall accuracy of all the test data and the arithmetic mean of the accuracy of 3 domains, respectively. Note that they are different because the capacities of different domains are not equal. The reported results are average over three repetitions of each run. All of the models are trained on Clipart, Infograph, and Quickdraw domains of DomainNet and tested on the other three domains. The title of each column indicates the name of the domain used as target. All the models are pretrained for 1000 epoches before finetuned on the labeled data. The best results of all methods are highlighted with the bold font.}
    \resizebox{0.85\textwidth}{!}{
    \begin{tabular}{c|ccc|cc|ccc|cc}
        \toprule
         & \multicolumn{5}{c|}{Label Fraction 1\%} & \multicolumn{5}{c}{Label Fraction 5\%} \\
         \midrule
         method & Painting & Real & Sketch & Overall & Avg. & Painting & Real & Sketch & Overall & Avg. \\
        \midrule
    
        MoCo V2 \cite{chen2020improved, he2020momentum}  & 11.38 & 14.97 & 15.28 & 14.04 & 13.88 & 20.80 & 24.91 & 21.44 & 23.06 & 22.39 \\
        SimCLR V2 \cite{chen2020big}  & \textbf{16.97} & 20.25 & 17.84 & 18.85 & 18.36 & 21.35 & 24.34 & 27.46 & 24.15 & 24.38 \\
        BYOL \cite{grill2020bootstrap}  & 5.00 & 8.47 & 4.42 & 6.68 & 5.96 & 9.78 & 10.73 & 3.97 & 9.09 & 8.16 \\
        AdCo \cite{hu2020adco}   & 11.13 & 16.53 & 17.19 & 15.16 & 14.95 & 19.97 & 24.31 & 24.19 & 23.08 & 22.82 \\
        FixMatch \cite{sohn2020fixmatch} &12.25 & 12.98 & 15.56 & 13.30 & 13.60 & \textbf{21.59} & 26.01 & 27.32 & 25.67 & 24.97\\
        ERM   & 6.68 & 6.97 & 7.25 & 6.94 & 6.96 & 7.45 & 6.08 & 5.00 & 6.24 & 6.18 \\
        \midrule
        DARL (ours)   & 14.45 & \textbf{21.68} & \textbf{21.30} & \textbf{19.59} & \textbf{19.14} & 21.09 & \textbf{30.51} & \textbf{28.49} & \textbf{27.48} & \textbf{28.19} \\
        
        \midrule
        & \multicolumn{5}{c|}{Label Fraction 10\%} & \multicolumn{5}{c}{Label Fraction 100\%} \\
         \midrule
         method & Painting & Real & Sketch & Overall & Avg. & Painting & Real & Sketch & Overall & Avg. \\
        \midrule

        MoCo V2  & 25.35 & 29.91 & 23.71 & 27.37 & 26.32 & 43.42 & 58.61 & 40.38 & 50.66 & 47.47 \\
        SimCLR V2  & 24.01 & 30.17 & 31.58 & 28.75 & 28.59 & 46.79 & 62.32 & 51.05 & 55.71 & 53.39 \\
        BYOL  & 9.50 & 10.38 & 4.45 & 8.92 & 8.11 & 34.02 & 46.48 & 24.82 & 38.59 & 35.11 \\
        AdCo  & 23.35 & 29.98 & 27.57 & 27.65 & 26.97 & 43.55 & 61.42 & 51.23 & 54.37 & 52.07 \\
        FixMatch & 25.15 & 32.39 & \textbf{33.18} & 30.54 & \textbf{30.24} & 44.76 & 55.15 & 54.93 & 52.22 & 51.62 \\
        ERM   & 9.90 & 9.19 & 5.12 & 8.56 & 8.07 & 31.50 & 40.21 & 24.01 & 34.48 & 31.91 \\
        \midrule
        DARL (ours)   & \textbf{25.90} & \textbf{33.29} & 30.77 & \textbf{30.72} & 29.99 & \textbf{49.64} & \textbf{63.77} & \textbf{54.31} & \textbf{57.91} & \textbf{55.91} \\
        \bottomrule
    \end{tabular}
    }
    \label{tab:res_all_correlated_domainnet}

\end{table*}

In Table \ref{tab:res_all_correlated_domainnet} we show the effectiveness of unsupervised learning methods for DG on setting \textsl{all correlated} with when source domains are Clipart, Inforgraph and Quickdraw and target domains are Painting, Real and Sketch, respectively.
Unsupervised training improves the generalization ability of models on the \textsl{All correlated UDG} setting. As explained in Section 4.1, this indicates the supervision from category labels of source domains is insufficient given that it can be considered as biased knowledge in target domains so that unsupervised pretraining is a effective approach for DG.

DARL consistently outperforms other unsupervised learning conterparts with all of the the split manners of pretraining/finetuning subsets, showing the superior of domain-irrelevant features against domain-relevant features when generalizing to novel domains.

One can consider the \textsl{all correlated} setting as a semi-supervised domain generalization setting. Thus we compare DARL with a SOTA semi-supervised method FixMatch\cite{sohn2020fixmatch}. DARL outperforms FixMatch on almost all of the subsets. Note that semi-supervised methods such as FixMatch can only be applied to deal with the \textsl{all correlated} setting but not other settings such as the \textsl{uncorelated} or the \textsl{domain correlated} setting. DARL shows much 
wider applicability compared to semi-supervised methods when the labeled data are heterogeneous and insufficient.  

Details of data split are shown in Table \ref{tab:count_all_correlated_domainnet_painting}, \ref{tab:count_all_correlated_domainnet_clipart}, \ref{tab:category_all_correlated_domainnet_painting}, and \ref{tab:category_all_correlated_domainnet_clipart}.

\subsection{Domain correlated UDG}
\begin{table*}[ht]
    \centering
    \caption{Results of the \textsl{domain correlated} setting on DomainNet. For details about the number of runs, meaning of column titles and fonts, see Table \ref{tab:res_all_correlated_domainnet}.}
    \resizebox{0.5\textwidth}{!}{
    \begin{tabular}{c|ccc|cc}
        \toprule
        & \multicolumn{5}{c}{Pretraining with data from 40 categories} \\
        \midrule
        method & Painting & Real & Sketch & Overall & Avg. \\
        \midrule
        MoCo V2 & 45.83 & 60.75 & 43.98 & 53.18 & 50.19 \\
        SimCLR V2 & \textbf{47.94} & 62.40 & 54.47 & 56.76 & 54.93 \\
        BYOL & 33.73 & 45.63 & 25.48 & 38.21 & 34.95 \\
        AdCo & 43.77 & 64.58 & 47.76 & 55.36 & 52.04 \\
        ERM & 31.92 & 41.58 & 24.10 & 35.32 & 32.53 \\
        \midrule
        DARL (ours)  & 47.82 & \textbf{65.07} & \textbf{56.90} & \textbf{58.61} & \textbf{56.60} \\
        \bottomrule
    \end{tabular}
    }
    \label{tab:res_domain_correlated_domainnet_clipart_40}
\end{table*}

We show that unsupervised learning improves models performance when the domains are re-split randomly in Table \ref{tab:res_domain_correlated_domainnet_clipart_40}. DARL achieves the best performance compared to other unsupervised methods, that indicates the superior of domain-irrelevant features against domain-relevant features when generalizing to novel domains.

\subsection{Uncorrelated UDG}
\begin{table*}[ht]
    \centering
    \caption{Results of the \textsl{uncorrelated} setting on CIFAR. Pre. and Fine. are short for pretraining and finetuning data. B, D, G, and S donate Brightness, Defocus blur, Gaussian noise and Snow, respectively. C, F, G, and S donates Contrast, Frost, Glass, Blur and Shot noise, respectively. 
    For details about the number of runs, meaning of column titles and fonts, see Table \ref{tab:res_all_correlated_domainnet}.}
    \resizebox{0.8\textwidth}{!}{
    \begin{tabular}{c|c|c|cccc|c}
        \toprule
         
         method & Pre. & Fine. & Elastic & Fog & Impulse Noise & Motion Blur & Avg. \\
        \midrule
    
        MoCo V2 & C, F, G, S & B, D, G, S & 76.33 & 70.45 & 74.73 & 72.57 & 73.52 \\
        SimCLR V2 & C, F, G, S & B, D, G, S & 77.91 & \textbf{72.38} & 76.00 & \textbf{73.46} & 74.94 \\
        BYOL & C, F, G, S & B, D, G, S & 58.32 & 50.88 & 54.34 & 56.95 & 55.13 \\
        AdCo & C, F, G, S & B, D, G, S & \textbf{79.67} & 71.05 & 74.35 & 72.17 & 74.31 \\
        ERM & C, F, G, S & B, D, G, S & 28.80 & 26.46 & 28.46 & 28.62 & 28.09 \\
        \midrule
        DARL (ours) & C, F, G, S & B, D, G, S & 78.98 & 71.52 & \textbf{76.65} & 73.19 & \textbf{75.09} \\
        \bottomrule
    \end{tabular}
    }
    \label{tab:res_uncorrelated_cifar}

\end{table*}

We show that unsupervised learning improves models performance when the domains are re-split randomly on the \textsl{uncorrelated UDG} setting in Table \ref{tab:res_uncorrelated_cifar}. DARL achieves the best performance compared to other unsueprvised methods, that indicates the superior of domain-irrelevant features against domain-relevant features when generalizing to novel domains.

\subsection{Finetuning with DG methods}

Here we further explore how unsupervised training affects the models trained with effective domain generalization methods. We show that DARL can be easily assembled with current DG methods to further improve the generalization ability. 

\textbf{Finetuning with DG methods} Table \ref{tab:res_dg_domainnet_clipart} shows how unsupervised pretraining methods benefit the generalization ability of empirical risk minimization (ERM) models since all the finetuning phase of these method can be considered as the training phase of ERM models. Here we further explore how unsupervised training affects the models trained with effective domain generalization methods. We report the results of state-of-the-art methods of domain generalization with unsupervised trained parameters as the initial state in Table \ref{tab:res_dg_domainnet_clipart}. 

With unsupervised learning, DG methods show stronger generalization ability. As a strong initialization, DARL learns a domain-irrelevant embedding space which maintains prior knowledge about distinguishing between domain-relevant and domain-irrelevant features.
Thus DARL can help DG methods including RSC and MMLD learn stronger generalization ability to unseen domains.  

If we consider DG methods as the backend of DARL, they achieve better performance compared to finetuning methods with ERM, as shown in Table \ref{tab:res_dg_domainnet_clipart}. Thus strong fully-supervised methods as the backend can further improve the performance of DARL. 

Then we compare DARL with DG methods that enlarge the available data space with augmentations of source domains. The generation based methods learns to generate samples while training, so that they can easily handle tasks with simple inputs, such as MNIST dataset. But when the structure of source data are complex, such as large scale real-world images, generation based methods usually require considerable computation cost and the effectiveness is harmed. DARL significantly outperforms generation based method such as M-ADA on the \textsl{domain correlated} setting, showing the superior of DARL for real-world DG tasks.

\begin{table*}[ht]
    \centering
    \caption{Results of state-of-the-art methods with different initialization methods under the \textsl{domain correlated} setting on DomainNet. For details about the number of runs, meaning of column titles and fonts, see Table \ref{tab:res_all_correlated_domainnet}.}
    \resizebox{0.6\textwidth}{!}{
    \begin{tabular}{c|ccc|cc}
        \toprule
        method & Painting & Real & Sketch & Overall & Avg. \\
        \midrule
        M-ADA \cite{qiao2020learning} & 31.24 & 44.96 & 25.51 & 37.17 & 33.90  \\
        RSC \cite{huang2020self} & 35.19 & 48.28 & 24.59 & 39.81 & 36.02 \\
        MMLD \cite{matsuura2020domain}  & 42.72 & 57.80 & 37.31 & 49.42 & 45.95 \\
        MoCo V2 + RSC   & 48.64 & 64.74 & 45.14 & 56.26 & 52.84 \\
        MoCo V2 + MMLD  & 51.57 & 65.75 & 50.52 & 58.70 & 55.94 \\
        \midrule
        DARL (ours) + RSC & \textbf{56.62} & \textbf{72.08} & \textbf{60.32} & \textbf{65.39} & \textbf{63.01} \\
        DARL (ours) + MMLD & 54.11 & 69.08 & 59.08 & 62.88 & 60.76 \\
        \bottomrule
    \end{tabular}
    }
    \label{tab:res_dg_domainnet_clipart}
\end{table*}

\section{Category correlated UDG}
We present the forth UDG setting, namely \textsl{category correlated UDG}, where unlabeled and labeled data share the same category space but not the domain space. This setting has practical significance, given we can consider any kind of data structure of the source other than category, such as the batch of medical data generation or the time real-world images are taken, as domain. Thus data from different domains can share the same category space. 

\begin{table*}[ht]
    \centering
    \caption{Results of the \textsl{category correlated} setting on CIFAR. Pre. and Fine. are short for pretraining and finetuning data. B, C, E, and G donate Brightness, Contrast, Elastic and Gaussian noise, respectively. F, G, M, and S donate Frost, Glass Blur Motion blur and Snow, respectively. 
    For details about the number of runs, meaning of column titles and fonts, see Table \ref{tab:res_all_correlated_domainnet}.}
    \begin{tabular}{c|c|c|cccc|c}
        \toprule
         
         method & Pre. & Fine. & Defocus Blur & Fog &  Impulse Noise & Shot Noise & Avg. \\
        \midrule
    
        MoCo V2 & B, C, E, G & F, G, M, S & 76.90 & 73.67 & 73.33 & 76.54 & 75.11 \\
        SimCLR V2 & B, C, E, G & F, G, M, S & 80.26 & 75.39 & 76.48 & \textbf{79.43} & 77.89 \\
        BYOL & B, C, E, G & F, G, M, S & 55.27 & 49.87 & 52.36 & 53.13 & 52.66 \\
        AdCo & B, C, E, G & F, G, M, S & 77.96 & 75.33 & 75.62 & 77.89 & 76.70 \\
        ERM & B, C, E, G & F, G, M, S & 26.53 & 23.34 & 25.67 & 27.94 & 25.87 \\
        \midrule
        DARL (ours) & B, C, E, G & F, G, M, S & \textbf{83.49} & \textbf{78.76} & \textbf{78.61} & 79.23 & \textbf{80.02} \\
        \bottomrule
    \end{tabular}
    \label{tab:res_category_correlated_cifar}

\end{table*}

We use CIFAR-10-C for this setting since the domain spaces of DomainNet(6) and PACS(4) are limited. To make sure the category spaces of unlabeled and labeled data are correlated, we randomly select 60\%, 20\% and 20\% of CIFAR-10 for pretraining, finetuning and test. We randomly select 4 specific
domains for pretraining, finetuning and test data, respectively. The details of data split are in Table \ref{tab:count_category_correlated}. 

Results are shown in Table \ref{tab:res_category_correlated_cifar}. Given pretraining, finetuning and test data share the same category space, the prior knowledge about category holds in the finetuning and inference phase. Thus unsupervised learning including MoVo V2, SimCLR V2 and BYOL achieve significant improvement for the \textsl{category correlated} setting. The embedding space DARL learns maintains more prior knowledge about the category compared to other unsupervised methods since the DARL prevents it from learning domain-related features. So DARL further improves the performance by 2.13\%.

\section{Ablation Study}
The contrastive loss in the proposed DARL relies largely on the similarity prediction network and domain specific negative sample sets. However, the domain specific negative sample sets are indispensable for DARL, thus we exploit the effect of similarity prediction network in this ablation study. 
A straightforward approach to hinder models from leveraging domain-related features to contrast negative pairs is splitting negative samples regarding to their domains and only calculating contrastive loss with negative pairs where two samples are from the same domain. Thus none of domain related features are considered in negative pairs. 

We present the results on \textit{domain correlated} setting with 100 categories in DomainNet for pretraining to conduct the ablation and show the results in Table \ref{tab:ablation}. Raw contrastive in the table refers to a model trained with contrastive loss without similarity prediction network and domain specific negative sample sets. DARL-st refers to the straightforward alternative of DARL without the similarity prediction network. DARL-st shows considerable improvement compared with raw contrastive loss, indicating that the adversarially generated domain specific negative samples are effective for helping the model learn category-discriminative features. Then the comparison between DARL-st and DARL further shows the effectiveness of the similarity prediction network, which learns more category-discriminative features from negative pairs from similar domains.

\begin{table*}[ht]
    \centering
    \caption{Results of the \textsl{domain correlated} setting on DomainNet for ablation study. For details about meaning of column titles and fonts, see Table \ref{tab:res_all_correlated_domainnet}.}
    \resizebox{0.6\textwidth}{!}{
    \begin{tabular}{c|ccc|cc}
        \toprule
         & \multicolumn{5}{c}{Pretraining with data from 100 categories} \\
         \midrule
         method & Clipart & Infograph & Quickdraw & Overall & Avg. \\
        \midrule
        ERM & 55.78  & 22.40 & 25.75 & 30.43 & 34.64 \\
        Raw contrastive & 75.32 & 32.96 & 38.17 & 43.67 & 48.82 \\
        DARL-st & 80.15 & 36.81 & 45.50 & 49.57 & 54.15 \\
        DARL & \textbf{82.28} & \textbf{40.60} & \textbf{47.68} & \textbf{52.19} & \textbf{56.85} \\

        \bottomrule
    \end{tabular}}
    \label{tab:ablation}
    \vspace{-10px}
\end{table*}

\begin{table*}[ht]
    \centering
    \caption{Data split details of \textsl{all correlated} setting when source domains are Painting, Real and Sketch. Pretrain, finetune, test indicate the number of available samples from corresponding domains in the pretraining, finetuning and testing phase, respectively.}
    \begin{tabular}{c|cccccc|c}
        \toprule
        phase & Painting & Real & Sketch & Clipart & Infograph & Quickdraw & All. \\
        \midrule
        pretrain & 5305 & 9896 & 3901 & 0 & 0 & 0 & 19102 \\
        finetune & 5305 & 9896 & 3901 & 0 & 0 & 0 & 19102 \\
        validation & 600 & 1111 & 444 & 0 & 0 & 0 & 2155 \\
        \midrule
        test & 0 & 0 & 0 & 3904 & 5348 & 10000 & 19252 \\
        \bottomrule
    \end{tabular}
    \label{tab:count_all_correlated_domainnet_painting}
\end{table*}

\begin{table*}[ht]
    \centering
    \caption{Data split details of \textsl{all correlated} setting when source domains are Clipart, Infograph and Quickdraw. Pretrain, finetune, test indicates the number of available samples from corresponding domains in the pretraining, finetuning and testing phase, respectively.}
    \begin{tabular}{c|cccccc|c}
        \toprule
        phase & Painting & Real & Sketch & Clipart & Infograph & Quickdraw & All. \\
        \midrule
        pretrain & 0 & 0 & 0 & 3504 & 4804 & 9000 & 17308 \\ 
        finetune & 0 & 0 & 0 & 3504 & 4804 & 9000 & 17308 \\ 
        validation & 0 & 0 & 0 & 400 & 544 & 1000 & 1944 \\
        \midrule
        test & 5905 & 11007 & 4345 & 0 & 0 & 0 & 21257 \\
        \bottomrule
    \end{tabular}
    \label{tab:count_all_correlated_domainnet_clipart}
\end{table*}

\begin{table*}[ht]
    \centering
    \caption{Data split details of \textsl{domain correlated} setting when source domains are Painting, Real and Sketch, while the number of class is 40. Pretrain, finetune, test indicate the number of available samples from corresponding domains in the pretraining, finetuning and testing phase, respectively.}
    \begin{tabular}{c|cccccc|c}
        \toprule
        phase & Painting & Real & Sketch & Clipart & Infograph & Quickdraw & All. \\
        \midrule
        pretrain & 9644 & 22421 & 10200 & 0 & 0 & 0 & 42265\\
        finetune & 4572 & 9726 & 4198 & 0 & 0 & 0 & 18496 \\
        validation & 516 & 1092 & 478 & 0 & 0 & 0 & 2086 \\
        \midrule
        test & 0 & 0 & 0 & 3471 & 5128 & 10000 & 18599 \\
        \bottomrule
    \end{tabular}
    \label{tab:count_domain_correlated_domainnet_painting_40}
\end{table*}

\begin{table*}[ht]
    \centering
    \caption{Data split details of \textsl{domain correlated} setting when source domains are Clipart, Infograph and Quickdraw, while the number of class is 40. Pretrain, finetune, test indicate the number of available samples from corresponding domains in the pretraining, finetuning and testing phase, respectively.}
    \begin{tabular}{c|cccccc|c}
        \toprule
        phase & Painting & Real & Sketch & Clipart & Infograph & Quickdraw & All. \\
        \midrule
        pretrain & 0 & 0 & 0 & 5304 & 5625 & 20000 & 30929 \\
        finetune & 0 & 0 & 0 & 3504 & 4804 & 9000 & 17308 \\
        validation & 0 & 0 & 0 & 400 & 544 & 1000 & 1944 \\
        \midrule
        test & 5905 & 11007 & 4345 & 0 & 0 & 0 & 21257 \\
        \bottomrule
    \end{tabular}
    \label{tab:count_domain_correlated_domainnet_clipart_40}
\end{table*}

\begin{table*}[ht]
    \centering
    \caption{Data split details of \textsl{domain correlated} setting when source domains are Painting, Real and Sketch, while the number of class is 100. Pretrain, finetune, test indicate the number of available samples from corresponding domains in the pretraining, finetuning and testing phase, respectively.}
    \begin{tabular}{c|cccccc|c}
        \toprule
        phase & Painting & Real & Sketch & Clipart & Infograph & Quickdraw & All. \\
        \midrule
        pretrain & 21213 & 49701 & 20868 & 0 & 0 & 0 & 91782 \\
        finetune & 4572 & 9726 & 4198 & 0 & 0 & 0 & 18496 \\
        validation & 516 & 1092 & 478 & 0 & 0 & 0 & 2086 \\
        \midrule
        test & 0 & 0 & 0 & 3471 & 5128 & 10000 & 18599 \\
        \bottomrule
    \end{tabular}
    \label{tab:count_domain_correlated_domainnet_painting_100}
\end{table*}

\begin{table*}[ht]
    \centering
    \caption{Data split details of \textsl{category correlated} setting. Pretrain, finetune, test indicate the number of available samples from corresponding domains in the pretraining, finetuning and testing phase, respectively.}
    \begin{tabular}{c|cccccc}
        \toprule
        phase & Brightness & Contrast & Elastic &  Gaussian Noise & Frost & Glass Blur  \\
        \midrule
        pretrain & 9000 & 9000 & 9000 & 9000 & 0 & 0  \\
        finetune & 0 & 0 & 0 & 0 & 3000 & 3000 \\
        \midrule
        test & 0 & 0 & 0 & 0 & 0 &  \\
        \midrule
        phase & Motion Blur & Snow & Defocus Blur & Fog & Impulse Noise & Shot Noise \\
        \midrule
        pretrain & 0 & 0 & 0 & 0 & 0 & 0 \\
        finetune & 3000 & 3000 & 0 & 0 & 0 & 0 \\
        \midrule
        test & 0 & 0 & 3000 & 3000 & 3000 & 3000 \\
        
        \bottomrule
    \end{tabular}
    \label{tab:count_category_correlated}
\end{table*}

\begin{table*}[ht]
    \centering
    \caption{Category space of pretraining, finetuning and test for \textsl{all correlated} setting when source domains are Painting, Real and Sketch.}
    \begin{tabularx}{\textwidth}{l|X}
        \toprule
        Phase & Available categories. \\
        \midrule
        pretrain, finetune, test & The\_Eiffel\_Tower, bee, bird, blueberry, broccoli, fish, flower, giraffe, grass, hamburger, hexagon, horse, sun, tiger, toaster, tornado, train, violin, watermelon, zigzag \\
        \bottomrule
    \end{tabularx}
    \label{tab:category_all_correlated_domainnet_painting}
\end{table*}

\begin{table*}[ht]
    \centering
    \caption{Category space of pretraining, finetuning and test for \textsl{all correlated} setting when source domains are Clipart, Infograph and Quickdraw.}
    \begin{tabularx}{\textwidth}{l|X}
        \toprule
        Phase & Available categories. \\
        \midrule
        pretrain, finetune, test & The\_Eiffel\_Tower, bee, bird, blueberry, broccoli, fish, flower, giraffe, grass, hamburger, hexagon, horse, sun, tiger, toaster, tornado, train, violin, watermelon, zigzag \\
        \bottomrule
    \end{tabularx}
    \label{tab:category_all_correlated_domainnet_clipart}
\end{table*}

\begin{table*}[ht]
    \centering
    \caption{Category space of pretraining, finetuning and test for \textsl{domain correlated} setting when source domains are Painting, Real and Sketch and the number of available category is 40.}
    \begin{tabularx}{\textwidth}{l|X}
        \toprule
        Phase & Available categories. \\
        \midrule
        pretrain & arm, backpack, basket, bear, beard, belt, bird, book, bridge, cat, cookie, couch, donut, drill, face, fan, finger, golf\_club, grass, helicopter, jacket, key, keyboard, lighthouse, mailbox, marker, mug, pencil, pizza, potato, shoe, shovel, sink, skyscraper, spoon, squirrel, sweater, telephone, tiger, train \\
        \midrule
        finetune, test & The\_Eiffel\_Tower, barn, bee, blueberry, broccoli, bus, butterfly, fish, giraffe, hamburger, hockey\_stick, sea turtle, spider, toaster, tornado, triangle, violin, watermelon, wine\_glass, zigzag \\
        \bottomrule
    \end{tabularx}
    \label{tab:category_domain_correlated_domainnet_painting_40}
\end{table*}

\begin{table*}[ht]
    \centering
    \caption{Category space of pretraining, finetuning and test for \textsl{domain correlated} setting when source domains are Painting, Real and Sketch and the number of available category is 100.}
    \begin{tabularx}{\textwidth}{l|X}
        \toprule
        Phase & Available categories. \\
        \midrule
        pretrain & The\_Great\_Wall\_of\_China, The\_Mona\_Lisa, apple, arm, asparagus, baseball\_bat, bathtub, belt, bicycle, binoculars, boomerang, bracelet, brain, bread, bucket, cake, calendar, candle, cannon, carrot, cat, ceiling\_fan, chair, circle, crayon, crocodile, cruise\_ship, cup, dolphin, door, dragon, dresser, elbow, elephant, eye, fire\_hydrant, flamingo, flashlight, flip\_flops, garden\_hose, hammer, harp, helmet, hot\_air\_balloon, hourglass, house, kangaroo, knife, leg, light\_bulb, lighter, line, mailbox, marker, microwave, monkey, moon, ocean, paintbrush, piano, pickup\_truck, pliers, pond, pool, potato, rabbit, rain, rake, rifle, river, rollerskates, sailboat, sandwich, saxophone, school\_bus, see\_saw, shoe, sink, snail, snorkel, snowflake, soccer\_ball, sock, spoon, square, string\_bean, swing\_set, table, telephone, tennis\_racquet, tent, toe, toothbrush, toothpaste, tree, trombone, umbrella, vase, waterslide, windmill \\
        \midrule
        finetune, test & The\_Eiffel\_Tower, barn, bee, blueberry, broccoli, bus, butterfly, fish, giraffe, hamburger, hockey\_stick, sea turtle, spider, toaster, tornado, triangle, violin, watermelon, wine\_glass, zigzag \\
        \bottomrule
    \end{tabularx}
    \label{tab:category_domain_correlated_domainnet_painting_100}
\end{table*}

\begin{table*}[ht]
    \centering
    \caption{Category space of pretraining, finetuning and test for \textsl{domain correlated} setting when source domains are Clipart, Infograph and Quickdraw.}
    \begin{tabularx}{\textwidth}{l|X}
        \toprule
        Phase & Available categories. \\
        \midrule
        pretrain & arm, backpack, basket, bear, beard, belt, bird, book, bridge, cat, cookie, couch, donut, drill, face, fan, finger, golf\_club, grass, helicopter, jacket, key, keyboard, lighthouse, mailbox, marker, mug, pencil, pizza, potato, shoe, shovel, sink, skyscraper, spoon, squirrel, sweater, telephone, tiger, train \\
        \midrule
        finetune, test & The\_Eiffel\_Tower, bee, bird, blueberry, broccoli, fish, flower, giraffe, grass, hamburger, hexagon, horse, sun, tiger, toaster, tornado, train, violin, watermelon, zigzag \\
        \bottomrule
    \end{tabularx}
    \label{tab:category_domain_correlated_domainnet_clipart_40}
\end{table*}

\end{document}